\pdfoutput=1
\documentclass[11pt]{article}

\usepackage{acl}

\usepackage{times}
\usepackage{latexsym}
\usepackage{graphicx}
\usepackage{xfrac}
\usepackage[T1]{fontenc}
\usepackage[utf8]{inputenc}
\usepackage{microtype}
\usepackage{amssymb}
\usepackage{mdframed}
\usepackage{lipsum}
\usepackage{booktabs}
\usepackage{comment}
\usepackage{amsmath}
\usepackage{xcolor}
\usepackage{soul}
\usepackage{multirow}
\usepackage{makecell}
\usepackage{aliascnt}
\usepackage{mathtools}
\usepackage{enumitem}

\definecolor{superlightred}{HTML}{F5F5F5}
\newmdtheoremenv[backgroundcolor=superlightred,
frametitlerulewidth=0pt,
innertopmargin=3pt,
innerrightmargin=3pt, 
innerleftmargin=3pt, 
innerbottommargin=3pt
]{hypothesis}{Hypothesis}
\newmdtheoremenv[backgroundcolor=superlightred,
frametitlerulewidth=0pt,
innertopmargin=3pt,
innerrightmargin=3pt, 
innerleftmargin=3pt, 
innerbottommargin=3pt
]{lemma}{Lemma}
\newmdtheoremenv[backgroundcolor=superlightred,
frametitlerulewidth=0pt,
innertopmargin=3pt,
innerrightmargin=3pt, 
innerleftmargin=3pt, 
innerbottommargin=3pt
]{proposition}{Proposition}
\newmdtheoremenv[backgroundcolor=superlightred,
frametitlerulewidth=0pt,
innertopmargin=3pt,
innerrightmargin=3pt, 
innerleftmargin=3pt, 
innerbottommargin=3pt
]{theorem}{Theorem}

\newenvironment{proofnamed}[1][\unskip]{\paragraph{#1:}}{\hfill$\square$}

\newcommand{\daniel}[1]{\textcolor{cyan}{[DK:#1]}}
\newcommand{\changedTwo}[1]{#1}
\newcommand{\changed}[1]{#1}
\newcommand{\sewon}[1]{\textcolor{magenta}{[SM:#1]}}

\newcommand{\kyle}[1]{\textcolor{red}{[KR:#1]}}
\newcommand{\tushar}[1]{\textcolor{orange}{[TK:#1]}}
\newcommand{\sameer}[1]{}

\newcommand{\name}{Prompt Waywardness}
\newcommand{\shortname}{Waywardness}
\newcommand{\set}[1]{\{#1\} }
\newcommand{\reals}{\mathbb{R}}
\newcommand{\prob}[1]{\mathbb{P}\left( #1 \right)}
\newcommand{\lm}[1]{\mathsf{LM}(#1)}

\newcommand{\loss}[1]{\mathsf{loss}(#1)}
\newcommand{\dist}{\mathsf{dist}}
\newcommand{\expec}[2]{\mathbb{E}_{#1}\left[#2 \right]}
\newcommand{\dataset}{\mathcal{D}}
\newcommand{\datasettr}{\mathcal{D}^{\text{train}}}
\newcommand{\datasette}{\mathcal{D}^{\text{test}}}

\DeclareMathOperator*{\argmin}{arg\,min}

\title{\textsc{\name}:\\ The Surprising Power and Risks of Continuous Prompts}
\title{\textsc{\name}:\\ The Power of Continuous Prompts and Its Risk of Discretized Interpretation}
\title{\textsc{\name}:\\ The Expressivity of Continuous Prompts and Its Risk of Discretized Interpretation}
\title{\textsc{\name}: The Expressivity of Continuous Prompts and the Risks of their Discretized Interpretation}
\title{\textsc{\name}: Challenges of Discretized Interpretation of Continuous Prompts}
\title{
\vspace*{-0.5in}
{{\small \hfill NAACL 2022}\\
\vspace*{.25in}} 
\textsc{\name}: The Curious Case of\\Discretized Interpretation of Continuous Prompts}

\author{
  Daniel Khashabi$^{\ddagger}$\ \ \;\; 
  Xinxi Lyu$^{\dagger}$\ \ \;\;   
  Sewon Min$^{\dagger}$
  \\
  \textbf{Lianhui Qin}$^{\dagger}$ \ \ \ \; 
  \textbf{Kyle Richardson}$^{\ddagger}$\ \ \ \; 
  \textbf{Sean Welleck}$^{\dagger \ddagger}$ 
  \\ 
  \textbf{Hannaneh Hajishirzi}$^{\dagger \ddagger}$  \
  \textbf{Tushar Khot}$^{\ddagger}$ \
  \textbf{Ashish Sabharwal}$^{\ddagger}$ \ 
  \textbf{Sameer Singh}$^{\ddagger \heartsuit}$ \ 
  \textbf{Yejin Choi}$^{\dagger \ddagger}$ 
  \\
  \\
 $^\dagger$University of Washington\ \ \; 
 $^\ddagger$Allen Institute for AI\ \ \; 
 $^\heartsuit$University of California-Irvine
}

\begin{document}
\maketitle
\begin{abstract}
Fine-tuning continuous prompts for target tasks has recently emerged as a compact alternative to full model fine-tuning. 
Motivated by these promising results, 
we investigate the feasibility of extracting a discrete (textual) interpretation of continuous prompts that is faithful to the problem they solve. 
In practice, we observe a ``\emph{wayward}'' behavior between the task solved by continuous prompts and the nearest neighbor discrete projections \changed{of these prompts:}
% \tushar{This sentence seems to imply that there is no correlation between the task being solved and discrete projection of the prompt. This is a valid point and may be we should talk about it in the paper too. The next sentence makes a different point: You can actually find continuous prompts that project to any arbitrary text. So rather than a ":", I would say "Moreover, we can find.."} 
\changed{One} can find continuous prompts that solve a task while being projected to an \emph{arbitrary} text (e.g., definition of a different or even a contradictory task) \changed{and simultaneously} being within a very small (2\%) margin of
the best continuous prompt of the same size for the task. 
We provide intuitions behind this odd and surprising behavior, as well as extensive empirical analyses quantifying the effect of design choices.
For instance, larger models exhibit higher waywardness, i.e, we can find prompts that more closely map to any arbitrary text with a smaller drop of accuracy. 
These findings have important implications relating to the difficulty of faithfully interpreting continuous prompts and their generalization across models and tasks, providing guidance for future progress in prompting language models. 
\end{abstract}

\section{Introduction}
Recent work has shown the surprising power of \emph{continuous prompts} to language models (LMs) for controlled generation and for solving a wide range of tasks~\cite{li2021prefix,lester2021power,min2021noisy}. 
% --- To date, we now have efficient algorithms that optimize (search for) \underline{continuous} prompts that solve tasks or lead to a particular generation. \sewon{I feel like this sentence contains the same information as the first sentence.} 
Despite these successes, the resulting continuous prompts are not easy to interpret~\cite{shin2020eliciting}. Is it possible to come up with meaningful discrete (textual) interpretations of continuous prompts, especially ones that provide a faithful explanation of the prompt's behavior?
% \sewon{I have a slight concern in using the term `generation' a lot since the datasets wise're using are fairly simple NLU datasets (classification).
% I think `controlled generation' is used to refer something different in the NLG community.}
% This model design paradigm has been touted as ``modular and space-efficient'' an alternative to fine-tuning LMs~\cite{li2021prefix} which involves changing the parameters of an architecture.
\begin{comment}
In parallel, manually-written \underline{discrete} prompts show interesting but complementary properties such as, 
transparency, good few-shot generalization~\cite{schick2021exploiting}, transfer across models~\cite{mishra2021reframing} and so on. 
A natural question is whether there is any correspondence between the two lines of work on continuous and discrete prompts. \tushar{Drawing parallels between these two style of prompts risks confusing the message. Esp since we are mostly focusing on \emph{discrete interpretation} of continuous prompts. I would suggest focusing on continuous prompt and motivate the need to interpret these prompts.}
% \sewon{Maybe we need to also mention somewhere about the whole point of prompt learning? e.g. training and saving large LMs for every task is extremely costly?}
\end{comment}

\begin{figure}
    \centering
    \includegraphics[scale=0.85,trim=0.65cm 0.6cm 0cm 0.4cm]{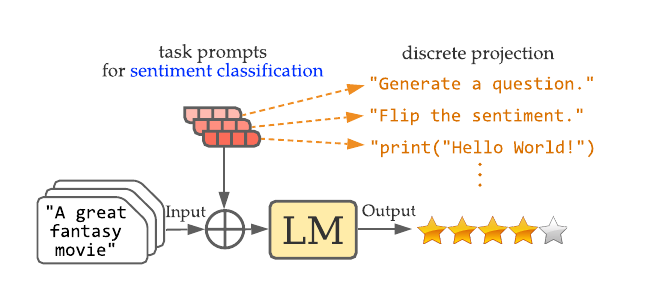}
    \caption{
        We show that one can find accurate continuous prompts (that do well on a given task, e.g., sentiment classification here) such that they can be projected to any \emph{arbitrary} text, such as the definition
        % we are able to find continuous prompt that perform pretty well on a task (e.g., sentiment classification), even though it is projected to an \emph{arbitrary} text, such as the definition 
        of a different task (e.g., generating a question) or even an irrelevant statement (e.g., a piece of code) --- suggesting a 
        % \name{} is the 
        disconnect between the outcome of continuous prompts and their discrete interpretations. 
        % \lianhui{It might be good to avoid using definite words like "always" (and perhaps also "any").}.
        % Similarly, a prompt touted as a fair resume selection model based on its its discrete interpretation might be carrying egregious biases.   
        % Examples of supervised and unsupervised prompting tasks. 
        % While investigating the possibility of projecting continuous prompts into meaningful language we observe \emph{\name}:   
        % The interpretation of continuous prompts for various tasks (supervised or otherwise) into meaningful discrete representation is impeded by \emph{\name}: 
        % there are infinite many continuous problems that lead to the desired output, many of which are projected to irrelevant and possibly, contradictory discrete representations. 
    }
    \label{fig:teaser}
\end{figure}

Towards addressing this question, we propose and investigate the \emph{\name} hypothesis (\S\ref{subsec:conjectures}), a surprising  disconnect between the intended behavior of continuous prompts and their nearest-neighbor discrete (language) representations.\footnote{
Nearest-neighbor projection via dot product has been previously used to study properties of continuous word embeddings~\cite{mikolov2013distributed,hashimoto2016word} and is commonly performed in the final layer of modern generative LMs~\cite{radford2019language,raffel2020exploring}. 
\label{footnote1:operator:generality}
} 
% Concretely, there are continuous prompt that assigns near perfect probability to the desired output (e.g., determining the sentiment of a sentence; Fig.\ref{fig:teaser}) while  being projected to \emph{any} given text.
In particular, we show that one can find continuous prompts that perform a desired task while, at the same time, project to \emph{any} given target text. This indicates that there is little correspondence between continuous prompts and their discrete interpretation. 
For instance, a continuous prompt that effectively solves the sentiment classification task in Fig.\ref{fig:teaser}, when projected onto discrete space, might appear as the definition of a different task (\emph{``flip the sentiment''}).
\changedTwo{
Intuitively, continuous prompt optimization is a highly non-convex problem with numerous local minima. More surprisingly, many of these prompts (e.g., those near embedded text) 
are very effective at solving a desired task.
}

% that it is possible to learn a continuous prompt that achieves a reasonable accuracy (near the ceiling accuracy using the same family of prompts), 
% such as their expressivity and the lack of one-to-one mapping that the many-to-one mapping of continuous to discrete space as well as the expressivity of deep neural networks .
% Suppose we find a continuous prompt that assigns high probabilities to \emph{``\_ are popular sites in Seattle.''} (Fig.\ref{fig:teaser}). 
% Consider the first example in  which involves an unsupervised discovery of prompts that 
% In an ideal world, the resultant prompt should be projected to a sensible discrete prompt (e.g., ``Its beautiful nature'').
% Alas, we observe that there are often many continuous prompts that assign high probability to the given context. 
% Consider a continuous prompt such that it's discrete projection states \emph{``Determine the sentiment''} (Fig.\ref{fig:teaser}; top) even though this prompt solves a completely different prompt (generates questions given an input sentence).  

We conduct extensive analyses showing \shortname{} on five classification datasets (\S\ref{sec:empirical}). 
Empirically, we find the existence of \emph{wayward} prompts --- prompts that solve each of these tasks while projecting to \emph{arbitrary} natural language text. 
To study a variety of projected text, we experiment with 60+
% 62 \lianhui{The number is a bit "magic" to me and wonder why is 62. We may not necessarily mention this specific number here?} \sewon{I sort of like mentioning the specific number, just to be concrete about the experiments as much as possible}\tushar{For an intro, maybe 60+ would sound better?}\ashish{60+ sounds good}
% different \ashish{drop "different"? unnecessary, unless you mean something other than non-identical} 
sentences, either discrete prompts from other tasks (from \citealt{mishra2021cross}) or random sentences from a large text corpus.
We observe that it is possible to 
% find prompts that reconstruct 95\% of a given discrete prompt (F1 token overlap)
find prompts that project to a given discrete prompt (token overlap 94\% F1) while scoring within 2\% accuracy of the best continuous prompts-based solution for the task. 
% In further analysis we quantify various parameters such as model size and prompt length. 
Further analysis shows that the effect of \shortname{} gets worse for larger models and longer prompts.
We explain this surprising behavior by \emph{relating it to} several structural properties of large language models (\S\ref{subsec:intuiton}). 

%\sewon{Maybe make it more precise? Accuracy of the prompt tuning that is optimized solely for accuracy} \daniel{rephrased it a bit so that it focuses on the experiments -- happier?}

We \changed{discuss several} social and research implications of prompt waywardness, to help guide future research on prompt based models (\S\ref{subsec:implications}). 
First and foremost, despite many promising attributes of continuous prompts, interpreting them is non-trivial and will require further research. 
In fact, careless interpretation of continuous prompts can result in vulnerabilities against malicious attacks concealed under the guise of benign discrete representation. 
% For the same reason, we hypothesize that continuous prompts will struggle in a several aspects of generalization. 
% For example, while discrete prompts trigger similar response on different models, continuous prompts might lead to vastly different outcomes on different models. \tushar{I think we can drop this conclusion from the intro. I feel lack of generalization of continuous prompts is a given even without waywardness.}
Further, the loose correspondence between continuous and discrete prompts poses a challenge for future research in \emph{differentiable interpretable-prompt optimization} -- optimization in search of human readable discrete prompts through the continuous space. 
Our work shows that continuous and discrete prompts, despite their seeming similarity, are quite different and the results from one may not always transfer to the other. 
% and enjoy different properties. 
% They should not be equated and as the results from one camp might not hold for the other one.
We hope these findings will motivate further innovations in the prompting literature for NLP models.

% for how the field should pursue the research on prompting literature. 
% Such attributes of continuous prompts make them an interesting technology and 
% Given such interesting attributes of continuous prompts, we hypothesize that this methodology will gain more popularity over fine-tuned models. 
% Therefore, it is necessary to foresee any dangers that may lay ahead. 

% Relatedly, the field has also explored the benefits of \emph{discrete} prompts. 
%  For example, hand-designed discrete prompts are shown to reduce the sample complexity of LMs~\cite{le2021many}.
%  They are also shown to support generalization across tasks~\cite{mishra2021cross,wei2021finetuned,sanh2021multitask} and models~\cite{mishra2021reframing}. However, unlike \emph{continuous} prompts, there are no known algorithms for directly optimizing (searching for) \emph{discrete} prompts. 
 % We want to understand their \daniel{their interface and how they are related to each other}

\section{Related Work}
\paragraph{Continuous prompts.}
There is  a line of work focused on tuning continuous prompts
\cite{li2021prefix,lester2021power,zhong2021factual,qin2021learning,zhou2021learning,zhong2021factual}. 
% Li and Liang (2021) applies a similar method for natural language generation and achieves comparable performance to fine-tuning while updating only 0.1\% of model parameters. 
These works present different approaches to discovering a continuous prompt (which is an array of real numbers) for addressing  an end task, though the interpretability of the resulting prompts remains an open question. 
% A recurring theme in this line of work is the strength of continuous prompt in results in strong, yet compact models---compared to conventional architecture fine-tuning approaches. 
% While we are excited by such findings, with the increasing popularity of continuous prompts it is also important to study their other properties such as the extent to which one can interpret them. 
% Motivated by the success of continuous prompt tuning, 
This paper investigates the feasibility of interpreting a learned continuous prompt and its connection to discrete prompts.

\paragraph{Discrete prompts.}
% In parallel to discrete prompts
The release of GPT-3~\cite{brown2020language} initiated a body of work on the emergent ability of LMs to follow  discrete natural language prompts.  
Consequently, several follow-up studies have used manually-designed discrete prompts for probing LMs~\cite{petroni2019language,jiang2020can},
improving LMs' few-shot ability~\cite{schick2021exploiting,gao2021making,le2021many}, and their zero-shot ability as well as transferability~\cite{mishra2021reframing,reynolds2021prompt}. 
% Despite the nice properties of discrete prompts,
Most importantly, discrete prompts have the advantages of being human-readable and thus easily interpretable
% While discrete prompts have clear advantages, in addition to being human-readable and thus easily interpretable \yejin{what are 'clear' advantages beyond human-readable and interpretable?}, 
though we do not have efficient and algorithmic ways of reconstructing them.
For example, \citet{shin2020eliciting}'s algorithm discovers discrete prompts, yet the results are not human readable.
   Prior work also finds that model performance is highly sensitive to small changes in wordings~\citep{mishra2021reframing} and that optimization over the discrete prompt space is non-trivial and often highly unstable.
Our findings here about the disconnect between continuous prompts and their discrete interpretation provides another perspective on the difficulty of discovering discrete prompts via continuous optimization algorithms that (directly or indirectly) leverage the continuous space (more discussion in \S\ref{subsec:implications}).

\section{\name}
\subsection{Preliminaries: Setup and Terminology}
\begin{figure}
    \centering
    \includegraphics[scale=0.75,trim=0.87cm 0.6cm 0cm 0.6cm,clip=true]{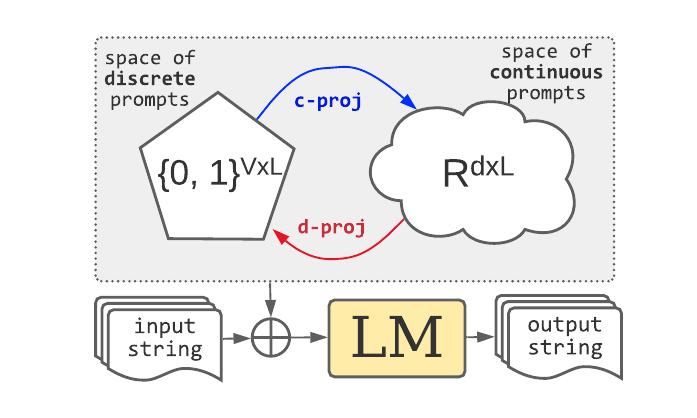}
    \caption{The general problem setup: 
    Similar to \citet{lester2021power}'s setup, each prompt (usually a continuous one) is appended to the given input and fed to a frozen language model. 
    % \daniel{need to clarify that sometimes what is appended is continuous and sometimes it is discrete; the figure does not clarify the setup for unsupervised task solving.} 
    }
    \label{fig:setup}
\end{figure}

We begin with some notation and the setup of our study, starting with the space of discrete and continuous prompts (Fig.\ref{fig:setup}). 
Let $p_d \in \set{0, 1}^{L \times V} $ denote a discrete prompt represented as an $L$-length sequence of  one-hot vectors over a lexicon of size $V$ (corners of a hyper-cube).  
Similarly, let $p_c \in \reals^{L\times d}$ denote a continuous prompt, represented as a $L$-length sequence of  $d$-dimensional real vectors. 

\paragraph{Projection operators.}
We define operators that project these two spaces to one another.
Define the $c$-projection as one that maps discrete inputs to a continuous space by  multiplying with a fixed (often pre-trained) embedding matrix\footnote{In our experiments we use the embedding matrix of the GPT2 family~\cite{radford2019language} which is used for both mapping input text to their embeddings as well as generating text in the output layer.} $E \in \reals^{V \times d}$:
\begin{equation}
    \label{eq:cproj}
    c\textnormal{-proj}(p_d) = p_d E \in \mathbb{R}^{L \times d}. 
\end{equation}
The $d$-projection maps the continuous inputs to nearest neighbor discrete elements, 
% : $d\textnormal{-proj}(p_c) \in \{0, 1\}^{L \times V},$ 
where for each position $l$ ($1 \leq l \leq L$), 
one of the possible (and perhaps most straightforward) methods for interpreting a continuous prompt is defined as a projection onto nearest neighbor representations~\cite{mikolov2013distributed,hashimoto2016word}: 
% \lianhui{Probably better to say this is “one of the possible (and perhaps most straightforward) methods for interpreting a continuous prompt”, since there is no previous work (I think) of interpreting continuous prompt (or if there is any, we may cite here), nor a widely accepted/used method for interpretation (so here we define one of the possible method, and base our study on it).} \lianhui{And indeed in the Intro and later part, we do have mentioned there are other ways of d-projection}\sameer{+1}: 
% \begin{equation}
%     \label{eq:dproj}
%     \mathrm{Onehot}_V(\argmax(p_c^l E^{\top})) \in \{0, 1\}^V,
% \end{equation}
% \tushar{I almost missed that this is the definition of d-proj. To be consistent with c-proj, can we also add an equation corresponding to d-proj?}\daniel{added an equation above. better?}\tushar{It helps a bit but I wanted more along the lines of:}
\begin{equation}
    \label{eq:dproj}
    d\textnormal{-proj}(p_c) = [\delta_1 \cdots \delta_l \cdots  \delta_L] \in \{0, 1\}^{L \times V},
\end{equation}
where $\delta_l$ is a one-hot vector corresponding to the word with the closest (highest dot product) embedding to the $l$-th position of continuous prompt $p_c$.
% \footnote{
% In practice this is usually implemented as a `softmax' function with a low temperature parameter to make the distribution peakier, similar to a one-hot vector~\cite{jan2017categorical}. 
% }\sean{why is softmax needed if it preserves max/min?}
% $\delta_l = \mathrm{Onehot}(\argmax_{w\in V}(p_c^l \cdot E_w))$. 
% \ashish{see comment on Slack: argmax doesn't quite make sense over a vector in $[0,1]^V$. Placing a comment to REVISIT THIS.}
% \ashish{I think argmax needs to be removed. what you need is the INDEX in a vector corresponding to the max value. that's not argmax. Just describe what $\delta_l$ is in words.}
% Here $p_c^l$ is the $l$-th element of $p_c$, 
% $E_w$ is the embedding corresponding to word $w$, and
% `$\mathrm{Onehot}$' maps a word $w$ name to its $V$-dimensional one-hot vector.
% % which $x$-th element is $1$ and other elements are $0$.

% is also defined similarly: 
% \begin{equation}
%     \label{eq:dproj}
%     d\textnormal{-proj}(p_c) =  \textnormal{softmax}(p_c \times E^{\top}/ \tau), 
% \end{equation}
% where $\tau$ is the temperature parameter, often set to a small number to ensure that the output is similar to one-hot vector. 
%
These projections are used in the first and last layer of virtually all modern LMs, such as GPT2.
% ~\cite{radford2019language}\sameer{weird to cite one. either cite a few, or none}.

\paragraph{Solving tasks with continuous prompts.}
Consider any machine learning model $M$ (typically a pre-trained model) that takes textual input $x$ and produces output $y$. Normally, the parameters of $M$ are learned so as to optimize behavior on a task with a dataset $\dataset = \set{(x, y)}$ of input/output pairs. In prompt tuning~\citep{lester2021power}, one freezes the parameters of $M$ and instead optimizes for a prompt $p$ that, when fed in conjunction with $x$, makes $M$ produce the desired output $y$. Thus, $p$ represents the only learnable parameters in this method. When $p$ is a discrete prompt with $k$ tokens, it can be simply concatenated with $x$, denoted $p + x$. In our study, $p$ will be a \emph{continuous} prompt (of length equal to the embedding of $k$ tokens). We will concatenate it with the embedding of the input $x$. For simplicity and with some abuse of notation, we use $p + x$ to denote concatenation in this continuous case as well.

One can quantify the amount of loss incurred when using a continuous prompt $p$ as follows: 
\begin{equation}
     \label{eq:main:obj}
     \ell(p; \dataset) = \expec{x, y\sim \dataset}{\loss{M(p + x), y}},
\end{equation}
% Here ``$+$'' denotes the concatenation of prompt $p$ to (the embedded representation of) input $x$. In continuous prompt tuning~\citep{lester2021power}, $p$ is the only learnable parameter and all parameters of the LM is frozen.
Minimizing this loss function (empirical risk minimization) over $p$ recovers a minimum risk continuous prompt for this dataset: 
% Suppose we are at the standard machine learning setup where the task is defined through observed $\dataset^{train}$ and held-out sets $\dataset^{test}$. 

\begin{equation}
    \label{eq:optimal:2}
    p_c^* =  \argmin_{p_c \in \reals^{L \times d}} \ell(p_c; \datasettr). 
\end{equation}
Given this prompt, its generalization to the test data can be measured in terms of the loss incurred on the test set: $\ell(p_c^*; \datasette)$.

%\sewon{I think we can shrink Section 2.1 to make it more concise if we are out of space---happy to try it when we are at that stage!}\tushar{At the least we should simplify this distinction between controlled generation and tasks. Just use the version with $\dataset$ where $\dataset$ can be a single example (controlled generation) or a task-specific dataset.} \sewon{I like this idea. At first I was confused what the difference is between two.}

\subsection{The \shortname{} Hypothesis}\label{subsec:conjectures}
How should one interpret the resultant continuous prompt $\tilde{p}_c$? 
Empirically, one can easily verify that such continuous prompts are not unique (e.g., random initializations lead to different outcomes). 
Additionally, the resultant prompts get projected to seemingly irrelevant discrete elements. 
% and they almost never get projected to human-readable discrete counter-part. 
% Taking this to an extreme, one can imagine that there exists a continuous prompt next to the projection of any discrete prompt, even if this discrete prompt is orthogonal to 
%  the content or even the the intent of the discrete prompt. \yejin{the later part of the previous sentence doesn't seem to read smoothly} 
Taking this to an extreme, we hypothesize that next to the continuous projection $c\text{-proj}(p_d)$ of any discrete prompt $p_d$, there exists a variety of continuous prompts $p_c$ that trigger responses from model $M$ that are orthogonal to the intentions described by the discrete prompt $p_d$. 
We formalize this idea as the following hypothesis, where $L \in \mathbb{N}$ is the length of the discrete target prompt, $M$ is a prompt-based model, and $\dataset$ is a dataset for a desired task:

% \begin{hypothesis}[\name]
%     \label{hypothesis}
%     % Let $L \in \mathbb{N}$, $M$ be a prompt-based model, and $\dataset$ be a dataset for a desired task.
%     % There exists a small $\Delta = \Delta(L, M, \dataset)$ such that for
%     For all $L,M,\dataset$,
%     there is a small $\Delta$ such that for
%     \underline{any} discrete target prompt $p_d$ with length $L$, \changed{there exists} a continuous prompt $\tilde{p}_c \in \reals^{L\times d}$ \changed{satisfying the following}: 
%     (1) $\tilde{p}_c$ is nearly as effective at making $M$ solve the task as the optimal continuous prompt (Eq.\ref{eq:optimal:2}), i.e.,  $|\ell(\tilde{p}_c; \dataset^{\mathit{test}})- \ell(p^*_c; \dataset^{\mathit{test}})| < \Delta$, yet 
%     % it is a strong-enough minimizer of the
%     (2) $\tilde{p}_c$ projects to $p_d$, i.e., $d$-proj$(\tilde{p}_c) = p_d$.
% \end{hypothesis}
% \sameer{hypothesis should have ``there exists'' or ``we can find'' attached to $p_c$ (with a small $\Delta$? there exists $\Delta$ is confusing..}

\begin{hypothesis}[\name]
    \label{hypothesis}
    For all $L,M,\dataset$, there is a small $\Delta$ such that for
    \underline{any} discrete target prompt $p_d$ with length $L$, there exists a continuous prompt $\tilde{p}_c \in \reals^{L\times d}$ such that: 
    \begin{enumerate}[itemsep=0mm,nolistsep]
        \item $\left| \ell(\tilde{p}_c; \dataset^{\mathit{test}})- \ell(p^*_c; \dataset^{\mathit{test}}) \right| < \Delta$, \ \ yet
        \item $d$-proj$(\tilde{p}_c) = p_d$.
    \end{enumerate}
\end{hypothesis}

In other words, $\tilde{p}_c$ is nearly as effective at making $M$ solve the task as the optimal continuous prompt (Eq.\ref{eq:optimal:2}), and yet it projects to $p_d$.
In this statement, $\Delta$ (prompt performance gap relative to the optimal prompt $p^*_c$) is a function of the prompt length $L$, the model $M$ (e.g., its embedding size and depth when $M$ is transformer based), and inherent properties of the target dataset $\dataset$.
The analysis in \S\ref{subsection:analyses} will provide an empirical estimate of this gap $\hat{\Delta}$ as a function of various parameters like model size and prompt length.

It is worth emphasizing that the hypothesis is stated for \emph{any} task and any set of discrete prompts, {\bf even if they are irrelevant or contradictory}.\footnote{While our focus is on the use of continuous prompts for solving datasets (one prompt shared among many instances), one can imagine applications of the same conjecture to special use cases such as controlled generation~\cite{dathathri2019plug} with one prompt per instance.}

\subsection{Finding Wayward Prompts}
\label{subsec:finding}

While the above hypothesis promises the \emph{existence} of $\tilde{p}_c$, it does not say how to discover them. We now discuss a practical method for their discovery. 

% \paragraph{Objective function.}
We learn a continuous prompt $p_c$ using a modification of the prompt tuning objective of \citet{lester2021power}. Our modification
%We modify the standard continuous optimization objective for controlled generalization (Eq.\ref{eq:optimal:1}) to
jointly minimizes the standard downstream task cross-entropy loss $\ell(.)$ for the task (Eq.\ref{eq:main:obj}) and a distance measure $\dist(.)$ between $p_c$ and the discrete target prompt $p_d \in \{0,1\}^{L \times V}$:
% \begin{equation}
% \begin{split}
% \begin{aligned}
\begin{align}
    \label{eq:modified:obj}
     \ell'(p_c; \dataset, \gamma) = & \ell(p_c; \dataset) + \gamma \, \dist(p_c, p_d) \\
    \tilde{p}_c = & \displaystyle \argmin_{p_c \in \reals^{L \times d}}  \ell'(p_c; \dataset, \gamma), 
\end{align}
% \tushar{the inputs to $\ell'$ don't seem to match up. How about changing the second one to $\ell'(p_c, D=\{(x,y)\}, \gamma)$?}
% \end{aligned}
% \end{split}
% \end{equation}
where $p_c$ is the only learnable parameter, and $\gamma$ is a hyperparameter.
%parameters controls the amount of weight put on optimizing prompts 

When $\gamma=0$, the modified objective is reduced to the standard objective (Eq.\ref{eq:optimal:2}), $\ell'(.) = \ell(.)$. 
We refer to this case and its resulting prompt $p^*_c$ as the {\em `unconstrained'} setting.  
A large value of $\gamma$ will make $p_c$ even closer (possibly identical) to $c\textnormal{-proj}(p_d)$ but lead to poor accuracy on a target dataset. 
Most of the experiments below are conducted via a range of $\gamma$ values to better understand the trade off between the two terms in the objective function. 
In practice, we find $\gamma=0.01$ to give a reasonable trade-off regardless of the target dataset and the choice of $p_d$.

There are at least two natural ways to define the distance measure $\dist(p_c, p_d)$ between a continuous prompt $p_c$ and a discrete target prompt $p_d$, by converting one so that both are in the same space:
\begin{align}
    \label{eq:dist-c} c\textnormal{-}\dist(p_c, p_d) = \frac{\| p_c  - c\textnormal{-proj}(p_d) \|^2_2}{L} \\
    \label{eq:dist-d} d\textnormal{-}\dist(p_c, p_d) = \mathrm{F1}\big(d\textnormal{-proj}(p_c), p_d \big)
\end{align}
The first of these places both $p_c$ and $p_d$ in the continuous space and computes the squared-L$^2$ norm, normalized by the prompt length. This is used in our training loss (Eq.\ref{eq:modified:obj}) implementation. The second places both in discrete space (text) and computes the standard word-level token overlap F1 score.\footnote{Ignoring punctuation marks and articles, and applying lemmatization.} This is used during our evaluation.

\section{Empirical Support of Waywardness}
\label{sec:empirical}

We empirically investigate the Prompt Waywardness hypothesis (\S\ref{subsec:conjectures}) using our modification (\S\ref{subsec:finding}) of the prompt tuning method from \citet{lester2021power}.
We show that given an {\bf arbitrary and irrelevant} discrete prompt $p_d$, it is possible to learn a continuous prompt that is mapped to $p_d$ while retaining its accuracy on a given dataset.\footnote{
Scripts needed to reproduce our results: 
\url{https://github.com/Alrope123/prompt-waywardness}
}

\subsection{Setup}
\label{subsec:setup}

\paragraph{Target tasks.} 
Following the setup of \citet{min2021noisy}, we select a diverse set of 5 classification datasets: SST-2~\citep{socher2013recursive}, SST-5~\citep{socher2013recursive}, AGNews~\citep{zhang2015character}, Subj~\citep{pang2004sentimental} and TREC~\citep{voorhees2000building}.
Statistics and the unconstrained accuracy of each dataset are provided in Table~\ref{tab:tasks}.
% \kyle{minor point: inlining table 1 now makes it look integrated within the last section, a bit confusing visually}

\newcommand{\aaa}{\mathcal{D}}

\begin{table}[ht]
    \centering
    \small
    \begin{tabular}{llcr}
        \toprule
        Dataset & Task &  $|C|$ & Acc \\
        \midrule
       SST-2   & Sentiment analysis (movie) & 2 & 92.4\\ 
        SST-5   & Sentiment analysis (movie) & 5 & 50.3\\ 
        AGNews & News classification (topic) & 4 & 88.1 \\ 
        Subj  &  Subjectivity classification & 2 & 90.5 \\
        TREC    &  Answer type classification & 6 & 88.0 \\  
        \bottomrule 
    \end{tabular}
    \caption{
        The collection of downstream tasks used in the experiments (\S\ref{subsec:setup}). 
        $|C|$ indicates the output size (number of classes); {\em Acc} indicates the unconstrained accuracy of a prompt tuning method~\citep{lester2021power} using GPT2 Large, as a reference point.
% \tushar{Add the prompt  length used here}\sewon{The prompt is not actually associated with the dataset here..? They are from Natural instructions or The PILE.}\tushar{Sorry this was a place-holder comment from a discussion, so not very clear. Basically what was the length of the continuous prompt used to achieve these unconstrained accuracies?}    
}
    \label{tab:tasks}
\end{table}

\paragraph{Discrete Target Projections.} 
We compile two sets of discrete target prompts: 
(1) 32 target prompts for solving tasks from Natural-Instructions\footnote{\url{https://instructions.apps.allenai.org}} dataset~\cite{mishra2021cross} that are distinct from and intentionally orthogonal to the end tasks considered here. 
These were chosen by excluding discrete target prompts that have high lexical overlap with other discrete prompts; this is because we found lexically similar prompts are often semantically similar even when written for different subtasks. 
(2) 30 random sentences from PILE,\footnote{\url{https://pile.eleuther.ai}} a large-scale, diverse text corpus used to pretrain GPT-J, the largest public causal language model~\citep{wang2021gpt}.
The sampled sentences were drawn from 
% We sampled sentences are drawn from the 
a Poisson distribution with $\lambda=14$, which makes the average length of the sentence to be consistent to those in Natural-Instructions.
These sentences are selected to have little or no token overlap with the true definition of the target tasks. 
% these selected discrete texts have almost none with the target prompts. 
See Table~\ref{tab:prompt_examples} for a few examples.

\newcommand{\gr}[1]{ \hspace{0.2cm}  \textcolor{gray}{(#1)} \hspace{0.27cm}}

\begin{table}[ht]
    \centering
    \small
    \begin{tabular}{llccc}
    \toprule
        \multirow{2}{*}{Data} & $p_d$ & Task Accuracy (\%)  & Prompt  \\
        & Source &  $\overbracket[0.2pt]{ \hat{\Delta} \; { \scriptsize  (\text{Acc}(p_c^*) \rightarrow \text{Acc}(\tilde{p}_c))}}$ & F1 (\%) \\
    \midrule
        \multirow{3}{*}{SST-2} 	& NI & 0.7 \gr{92.4 $\rightarrow$ 91.8} & 99.0\\
        & PILE & 0.5 \gr{92.5 $\rightarrow$ 92.0}  & 97.1\\
        & Avg & 0.6  \gr{92.4 $\rightarrow$ 91.9}		& 98.1 \\
    \cmidrule(lr){1-4}
        \multirow{3}{*}{SST-5} & NI	& 3.3 \gr{50.2 $\rightarrow$ 48.5}  & 95.9\\
        & PILE & 0.7 \gr{50.5 $\rightarrow$ 50.2}  & 92.4\\
        & Avg &  2.0 \gr{50.3 $\rightarrow$ 49.3} 	& 94.2	\\
    \cmidrule(lr){1-4}
        \multirow{3}{*}{AGNews} & NI	& 1.6 \gr{88.0 $\rightarrow$ 86.6}  & 97.4\\
        & PILE & -0.1 \gr{88.1 $\rightarrow$ 88.2}  & 97.5\\
        & Avg &  0.8 \gr{88.1 $\rightarrow$ 87.3}	& 97.4 \\
    \cmidrule(lr){1-4}
        \multirow{3}{*}{Subj} & NI	& 2.0 \gr{91.3 $\rightarrow$ 89.5}   & 97.3\\
        & PILE &  0.9 \gr{89.6 $\rightarrow$ 88.8}   & 94.4\\
        & Avg & 1.5 \gr{90.5 $\rightarrow$ 89.2}	&  95.9 \\
    \cmidrule(lr){1-4}
        \multirow{3}{*}{TREC} & NI	& 3.3 \gr{87.5 $\rightarrow$ 84.7}   & 86.5\\
        & PILE & 1.2 \gr{88.5 $\rightarrow$ 87.5}   & 85.6\\
        & Avg & 2.3 \gr{88.0 $\rightarrow$ 86.0}	& 86.1 \\
    \bottomrule 
    \end{tabular}
    \caption{
         Main Results: Accuracy of solving five classification datasets, 
         in an unconstrained setting ($p_c^*$)
        %  \sewon{Does it mean unconstrained accuracy? Why not call it unconstrained accuracy, as it is already defined?}
         vs.\ constrained by the projection to various {\bf irrelevant} pieces of text ($\tilde{p}_c$). Optimization is done using $\gamma = 0.01$ in the objective function (Eq.\ref{eq:modified:obj}).
         $\hat{\Delta}$ indicates the relative accuracy drop (in \%) from unconstrained accuracy.
         Each reported score ({\em Accuracy} and {\em Prompt F1}) are the average over 62 discrete target prompts and 3 random seeds.
         \textbf{Overall, it is possible to achieve $\boldsymbol{\geq 94\%}$  prompt F1 with under $\boldsymbol{2\%}$ drop in accuracy.}
    }
    \label{tab:main_result}
\end{table}

%\begin{figure*}[!th]
%    \centering
%    \resizebox{2.1\columnwidth}{!}{
%    \includegraphics[trim=0.0cm 0.8cm 0cm 0.5cm]{figure/main_result.pdf}
%    %\rule{6cm}{1cm}
%    }
%    \caption{
%        Accuracy of solving five classification datasets, projected to various discrete representation using three values of $\gamma$ ($0.01, 0.005, 0.003$) in the objective function (Eq.\ref{eq:modified:obj}).
%        For comparison, the dotted lines indicates an accuracy of a model with no projection (unconstrained accuracy, $\gamma = 0$).
%        Across all the downstream task we find that it is possible to learn a soft prompt $p_c$  which projected to an irrelevant discrete prompt, yet it retains the task accuracy. 
%        Each reported score ({\em Accuracy} and {\em Prompt F1}) are the average over 183 experiments (62 discrete prompts and 3 random %seeds).
%    }
%    \label{fig:main_result}
%\end{figure*}

\definecolor{darkred}{HTML}{C41E3A} %A52A2A}
\newcommand{\co}[1]{{\color{darkred}#1}}
\newcommand{\myskip}[1]{}

\myskip{
\begin{table*}[t]
    \centering
    \footnotesize
    \resizebox{\textwidth}{!}{  
    \begin{tabular}{l @{\hspace{-1.5em}} r}
        \toprule
        $d\textnormal{-proj}(p_c)$ & \emph{Prompt F1} \\
        \midrule
        \multicolumn{2}{c}{$p_d$: Please use your knowledge to determine if the following sentence is subjective or objective. } \\
        Please use your knowledge\co{,} determine if the following sentence is subjective or objective.
        & 96.0 \\
        Please use/ knowledge\co{.} determine if the following sentence is subjective or objective.
        & 91.7 \\
        Please use\co{,} knowledge\co{.} determine If the following sentence is \co{on} or objective. & 83.3 \\
    \midrule      
        \multicolumn{2}{c}{$p_d$: "So we might see you around?" she asked. The chill in her voice almost made him shudder. } \\
        "So we might see you around?" she asked. The chill in her voice almost made him shudder. & 100.0\\
        "So we might see you around?" she asked. The chill\co{.} her voice almost made him shudder. & 96.8 \\
        "So we might see you around?" she asked. The chill in her voice \co{subjective} made him shudder. & 93.8 \\
        "So we might see you around?" she asked. The chill in\co{?????-?????-} voice \co{okay} made him shudder. & 90.3 \\
        "So we might see \co{is} around?" she asked. The chill in her voice almost made him \co{subjective}. & 87.5 \\
   \midrule     
        \multicolumn{2}{c}{$p_d$: "I wanted to help. I couldn't stand by and let her treat you like that." } \\
        "I wanted to help. I couldn't stand by and let her treat you like that."
        & 100.0 \\
        "I wanted to help. I couldn't stand by and let \co{A} treat you like that."
        & 96.6
        \\
        "I wanted to help. I couldn't stand by and let her treat you like \co{TheNitromeFan}."
        & 93.3 \\
        "I wanted to help. I couldn't stand by and let her treat \co{is subjective} that."
        & 86.7 \\
    \midrule
        \multicolumn{2}{c}{$p_d$: 
        3.  \$f(Z):=\textbackslash~sum\_\{k=0\}\^~\textbackslash~infty a\_k Z\^~k\$ is a free holomorphic function on the open operatorial unit \$1\$-ball.
        } \\
        3.  \$f(Z):=\textbackslash~sum\_\{k=0\}\^~\textbackslash~infty a\_k Z\^~k\$ is a free holomorphic function on the open operatorial unit \$1\$-ball.
        & 100.0\\
        3. \$f(Z):=\textbackslash~sum\_\{k=0\}\^~\textbackslash~infty a\_k Z\^~k\$ is a free holomorphic function on \co{Boehner} open operatorial unit \$1\$-ball.
        & 96.3 \\
        3. \$f(Z):=\textbackslash~sum\_\{k=0\}\^~\textbackslash~infty a\_k Z\^~k\$ is a free \co{subjective}? function on the open operatorial unit \$1\$-ball.
        & 92.3 \\
        3. \co{????????} \$f(\co{Entity}):=\textbackslash~sum\_\{k=0\}\^~\textbackslash~infty a\_k Z\^~k\$ is a free holomorphic function on the open operatorial unit \$1\$-ball.
        & 82.8 \\
    \bottomrule 
    \end{tabular}
    }
    \caption{Examples of the target prompts $p_d$ and their reconstructed\kyle{reconstructions} via  $d\textnormal{-proj}(p_c)$ for different ranges of prompt F1 scores. 
    The first $p_d$ is from Natural-Instructions; other three are sampled from The PILE.
    The mismatches with the original prompt are \co{color-coded}. 
    \tushar{What are the associated tasks/dataset for each prompt?}
    \sewon{Looks like the first one is from Subj, and other three are not from tasks/datasets but are randomly sampled from The PILE.}\tushar{Hmm. Just to be clear I am not asking for the source of the prompts. I was interested in what was the task for which these prompts were learned. E.g. the almost meaningless equation in the bottom row is solving the task of "...".}
    \kyle{+1, having that somewhere would be helpful.}}
    \label{tab:prompt_examples}
\end{table*}
}

\begin{table*}[t]
    \centering
    \footnotesize
    \resizebox{\textwidth}{!}{  
    \begin{tabular}{l @{\hspace{-1.5em}} rr}
        \toprule
        $d\textnormal{-proj}(p_c)$ & \emph{Prompt F1} & \emph{Acc}($\tilde{p}_c$) \\
    \midrule
        \multicolumn{3}{l}{
        Task: AGNews \quad
        $p_d$: Write down the conclusion you can reach by combining the given Fact 1 and Fact 2.
        } \\
    \cmidrule(lr){1-3}
        Write down the conclusion you can reach by combining the given Fact 1 and Fact 2. & 100.0 & 89.2 \\
        Write down the conclusion you can reach by combining the given Fact 1\co{.} Fact 2. & 96.3 & 88.1 \\
        Write down the conclusion you can reach by combining the given Fact 1 \co{Category} Fact 2. & 92.9 & 89.0 \\
        Write \co{Messi in} conclusion you can reach by combining the given Fact 1 and Fact 2. & 89.7 & 88.8 \\
    \midrule
        \multicolumn{3}{l}{
        Task: SST-5 \quad
        $p_d$: ``If they have other interests and aims in life it does not necessarily follow that they are passive sheep.''
        } \\
    \cmidrule(lr){1-3}
        ``If they have other interests and aims in life it does not necessarily follow that they are passive sheep.'' & 100.0 & 51.2 \\
        ``If they have other interests and aims in life it does not necessarily follow that they are \co{terrible} sheep.'' & 94.7 & 53.6 \\
        ``If they have other interests and aims in life it does not necessarily follow that they are \co{terrible GoPro}.'' & 89.5 & 52.3 \\
    \midrule
        \multicolumn{3}{l}{
        Task: SST-5 \quad
        $p_d$: int clamp(int val, int min\_val, int max\_val) \{ return std::max(min\_val, std::min(max\_val, val)); \}
        } \\
    \cmidrule(lr){1-3}
        int clamp(int val, int min\_val, int max\_val) \{ return std::max(min\_val, std::min(max\_val, val)); \}
        & 100.0 & 50.5 \\
        int clamp(int val, int min\_val, int max\_val) \{ return std::max(min\_val, std::min(max\_val \co{terrible} val)); \}
        & 95.7 & 52.0 \\
        int clamp(int val, int min\_val, int max\_val) \{ return std::max(min\_val, std::min(max\_val \co{terrible} val)); \co{This}\}
        & 91.7 & 53.3 \\
    \bottomrule 
    \end{tabular}
    }
    \caption{Examples of the target prompts $p_d$ and their reconstructions via  $d\textnormal{-proj}(p_c)$ for different ranges of prompt F1 scores. 
    The first $p_d$ is from Natural-Instructions; the rest two are sampled from The PILE.
    The mismatches with the original prompt are \co{color-coded}. }
    \label{tab:prompt_examples}
\end{table*}

\paragraph{Evaluation metrics.}
For all experiments, we report two metrics: (1) the task \emph{accuracy}\footnote{
We did not consider alternatives like Macro-F1 because all datasets are roughly balanced across different classes.
%\daniel{can we say that the tasks are semi-balanced and hence, it is okay to use accuracy?}
} as well as (2) \emph{prompt F1}, the word-level token overlap F1 score
% between the desired discrete prompt $p_d$ and a projection of the learned continuous prompt to a discrete space $d\textnormal{-proj}(p_c)$
computed as in Eq.\ref{eq:dist-d},
since it easy to interpret and is commonly used for evaluating the textual output of models~\cite{rajpurkar2016squad}. 

\paragraph{Models.} 
For evaluation, we use GPT2~\cite{radford2019language} an auto-regressive LM which has extensively been used in many NLP applications. 
Unless otherwise specified, we use a `large' variant consisting of 774M parameters. 

\paragraph{Implementation details.}
We use a batch size of 8, learning rate 0.01, and 2000 training steps.
When experimenting with a discrete target prompt $p_d$, we initialize the search for continuous prompts (both $\tilde{p}_c$ and $p^*_c$) using $c\textnormal{-proj}(p_d)$.\footnote{
While this is different from prior work~\citep{lester2021power,min2021noisy} that uses a random subset of the top-5000 vocabs, we find no meaningful differences in an unconstrained accuracy between two initialization methods.
}
For all experiments, report accuracy averaged over three random seeds.

\subsection{Main Results}
\label{subsec:result}

For each of the 5 tasks $T$ and for each of the 62 discrete target prompts $p_d$, we use the objective in Eq.\ref{eq:modified:obj} to find a prompt $\tilde{p}_c$ such that it solves $T$ with a high accuracy while, at the same time, having a discrete projection that is close to $p_d$.  
For comparison, we also train unconstrained prompts ${p}^*_c$ ($\gamma=0.0$) which solve task $T$ without any projection constraint. 
To ensure a fair comparison between $\tilde{p}_c$ and ${p}^*_c$, we ensure that they have the same size $L$. 
In other words, for each $\tilde{p}_c$ (that has the same length as $p_d$), we train another ${p}^*_c$ with the same length.  
We denote the relative accuracy drop from ${p}^*_c$ to $\tilde{p}_c$ as $\hat{\Delta}$.

% \yejin{while the relatively small performance drop is surprising, it's also interesting to note that the performance almost always drop, instead of increasing, which seems to suggest that the original prompt is still superior over the nonsensical prompt. we should say something more precise about this}\ashish{Is there a confusion here that the ``original prompt" is discrete? Perhaps clarify: Note that since we are comparing to the optimal \emph{continuous} prompt $p^*_c$, using $\tilde{p}_c$ can only decrease performance (barring suboptimal search for $p^*_c$).} 
% \ashish{Alternatively (or in addition), as YC suggested, just take the embedding of the ``true'' discrete prompt (instead of optimizing for something close to it) and see how that performs compared to $\tilde{p}_c$ from this section; it will likely perform much worse, as it's not optimized for the task.}

% \daniel{Note that unconstrained accuracy also differ across $p_d$ because the continuous prompts are initialized from $p_d$.}

Table~\ref{tab:main_result} summarizes the results.  
Across all datasets, we find that it is possible to learn a continuous prompt
$p_c$ whose discrete projection is very close to $p_d$ and mostly retains the task accuracy. 
There is a trade-off between the task accuracy and prompt F1, which can be controlled by the choice of $\gamma$ (more extensive ablations in the forthcoming paragraphs (\S\ref{subsection:analyses})). Overall, with $\gamma=0.01$, it is possible to achieve $\geq 94\%$ prompt F1 with under $2\%$ relative drop in task accuracy. 
% As an exception, the prompt has lower prompt F1 on the TREC dataset: with 2.3\% relative drop in accuracy, prompt F1 is $86\%$ which is relatively low compared to other datasets.
The only outlier is the TREC dataset where we achieved a prompt F1 score of $86\%$ for a $\hat{\Delta}=2.3\%$ relative drop in accuracy.
% This might is likely because learning an effective continuous prompt on TREC is relatively hard due to its inherent differences with language modeling, as discussed by \citet{min2021noisy}.
This might be due to the difficulty of  learning effective prompts on TREC
% because of its differences with language modeling, 
(also discussed by \citet{min2021noisy}).

Example prompts with varying values of prompt F1 scores are shown in Table~\ref{tab:prompt_examples}.
A  prompt F1 $\geq 94\%$ generally indicates one word mismatch with almost no semantically meaningful difference. 
% ; $\geq 90\%$ prompt F1 indicates around 1 word flipped.
% Generally, $\geq 94\%$  prompt F1 indicates no semantically meaningful difference; $\geq 90\%$ prompt F1 indicates around 1 word flipped.

\begin{figure*}[!th]
    \centering
    \resizebox{2.11\columnwidth}{!}{\includegraphics[trim=0cm 0.8cm 0cm 0.2cm]{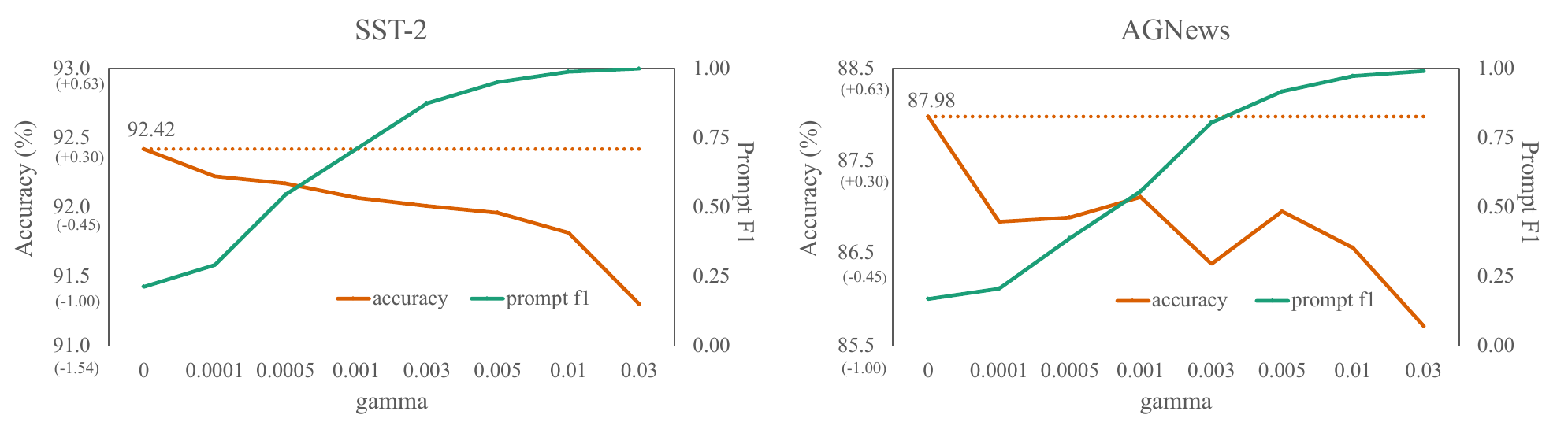}}
    \caption{The effect of $\gamma$ on SST-2 and AGNews.
    {\em Accuracy} is the average over 32 discrete target prompts from Natural Instructions and 3 random seeds.
    A dotted line indicates unconstrained accuracy $p^*_c$ (same as when $\gamma=0$).
    Numbers inside parentheses in the y-axis indicate relative drop in accuracy against unconstrained accuracy.
    \textbf{There is a clear trade-off between the task accuracy and the prompt F1.}
    }
    \label{fig:ablation_gamma}
\end{figure*}

%\begin{figure*}[!th]
%    \centering
%    \resizebox{2.1\columnwidth}{!}{
%    \includegraphics{figure/effect_of_prompt_length.pdf}
%    %\rule{6cm}{1cm}
%    }
%    \caption{The effect of the length of the prompt ($L$) on SST-2 and AGNews.
%    {\em Accuracy} is the average over 30 discrete prompts from the PILE and 3 random seeds.
%    A dotted line indicates unconstrained accuracy (same as when $\gamma=0$).
%    }
%    \label{fig:ablation_prompt_len}
%\end{figure*}

%\begin{figure}[!th]
%    \centering
%    \resizebox{\columnwidth}{!}{
%    \includegraphics{figure/effect_of_lm_size.pdf}
%    %\rule{6cm}{1cm}
%    }
%    \caption{The effect of the size of the model---distilled, small, medium, large, and XL---on SST-2.
%    {\em Accuracy} is the average over 32 discrete prompts from Natural Instructions and 3 random seeds.
%    A dotted line indicates unconstrained accuracy (same as %when $\gamma=0$).
%    }
%    \label{fig:ablation_model_size}
%\end{figure}

\subsection{Further Analysis}
\label{subsection:analyses}

\paragraph{Effect of Gamma.}
Fig.~\ref{fig:ablation_gamma} shows the trade-off between task accuracy and the prompt F1 when varying $\gamma$ from 0 to 0.03. As $\gamma$ increases, the task accuracy goes down while the prompt F1 increases. The drop in task accuracy is relatively minor---it is possible to learn a continuous prompt for which prompt F1 is near 1.00 and the accuracy drop relative to the unconstrained accuracy is less than 1\%.

\begin{figure}[ht]
    \centering
    \includegraphics[scale=.30,trim=2.8cm 0.0cm 0.0cm 0.0cm]{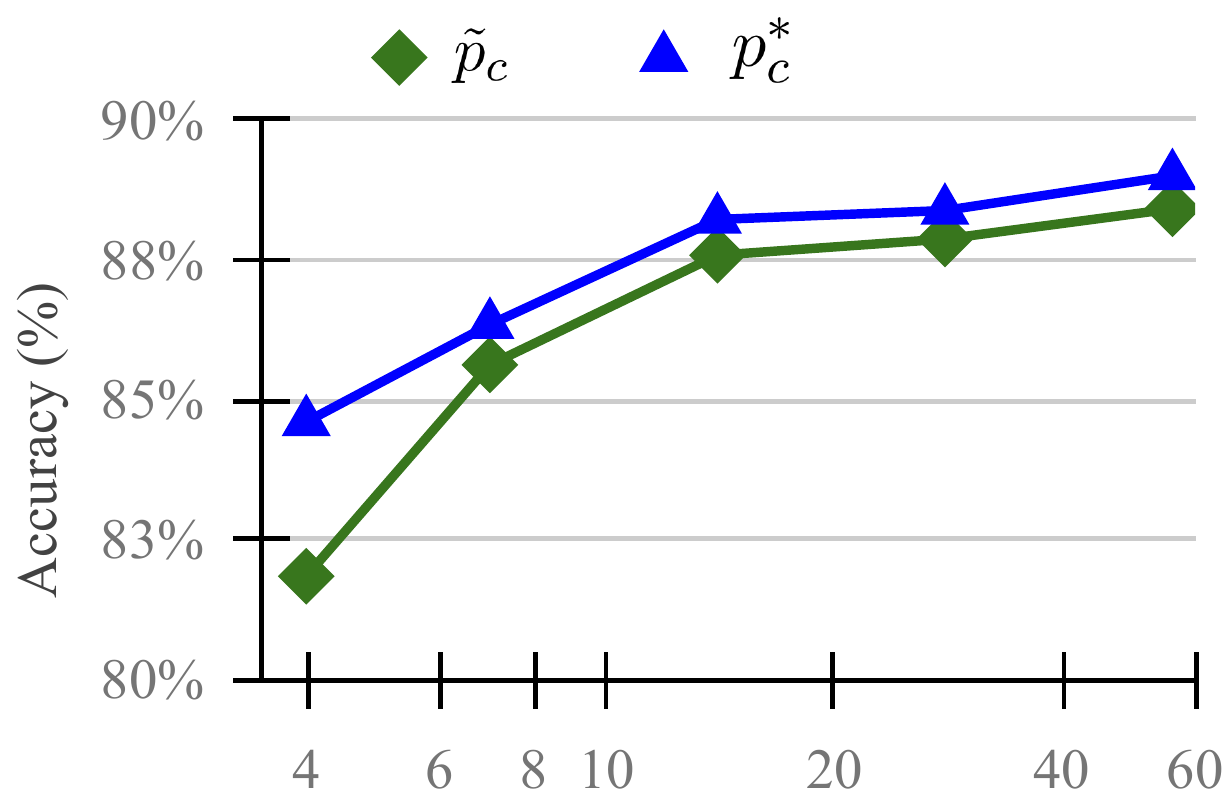}
    \includegraphics[scale=.65,trim=1.3cm 0cm 0cm 0cm]{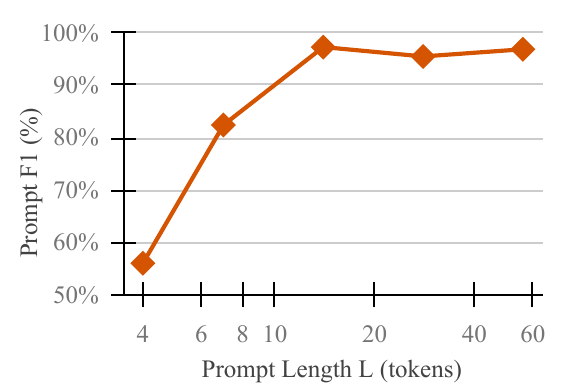}
    \caption{
    The effect of the length of the prompt ($L$) on AGNews.
    %{\em Relative Accuracy Drop} (green) is calculated against unconstrained accuracy, and is averaged over 32 discrete prompts from Natural Instructions and 3 random seeds ($\gamma=0.01$ used).
    Each point computed via the average over 32 discrete target prompts from Natural Instructions and 3 random seeds ($\gamma=0.01$ used).
    The corresponding prompt F1 is reported as a orange line.
    The accuracy of $p_c^*$ and $\tilde{p}_c$ increase as a function of prompt length, however, the gap between them tends to decrease. 
    \textbf{The relative accuracy drop is marginal when $L$ is not too small (e.g., 7 or larger).}
    % \tushar{You are probably going to fix this, but the legend for the blue line should be $p*$. Also minor nit-pick, can we make the $c$ an actual sub-script.}
    }
    \label{fig:length}
\end{figure}

\begin{figure}[ht]
    \centering
    \includegraphics[scale=.30,trim=2.8cm 0.0cm 0cm 0.0cm]{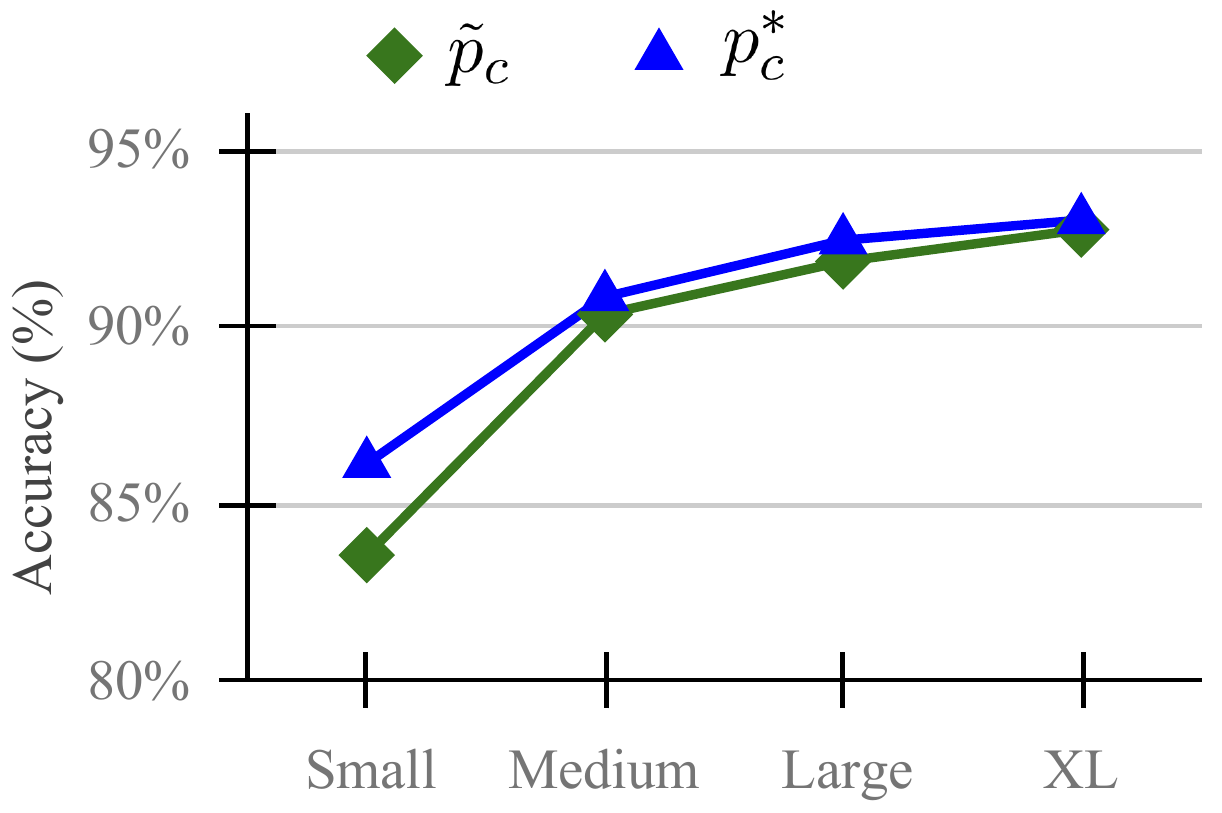}
    \caption{
    The effect of the size of the model---small, medium, large, and XL---on SST-2.
    Each point in the experiment is computed by averaging over 30 experiments each with a different discrete target prompt from PILE and 3 random seeds.
    We vary $\gamma=\{0.01, 0.005, 0.003\}$ and choose the one for which prompt F1 is larger than 0.98.
    \textbf{The relative accuracy drop (gap between the two trends) decreases as models become larger.}
    % \sameer{latex in legend is WIP, not final?}
    }
    \label{fig:model:size}
\end{figure}

\paragraph{Effect of Prompt Length ($L$).}
We randomly sample sentences from The PILE with a constraint that its length must be $L$ (chosen from $\{4, 7, 14, 28, 56\}$). The left and the middle parts of Fig.~\ref{fig:length} illustrate the results. We find that when $L$ is very small (e.g., 4) it is relatively difficult to learn a continuous prompt $p_c$ that is close to $p_d$ (F1<60\%)  while retaining the task accuracy.
This is likely because the prompt being too short significantly hurts the expressivity of the prompt.
Nonetheless, when $L$ is reasonably larger, e.g., $14$ (the average length of in Natural Instructions) or longer, all cases lead to a continuous prompt with near 1.0 prompt F1 and little accuracy drop.

\paragraph{Effect of Model Size.}
We vary the size of the GPT2 models---small, medium, large, and XL---with
%82M,
124M, 355M, 774M, and 1.5B parameters, respectively.
Figure~\ref{fig:model:size} (right) reports the result on SST-2. We find that (1) across different sizes of the LM, our findings in learning continuous prompts with the prompt F1 of near 1.0 and little drop in the accuracy generally hold, and (2) in particular, the drop in accuracy is more negligible with larger LMs (0.2\% with XL, 0.5--0.7\% with medium and large, 1.2\% with small).

\begin{table}[ht]
    \centering
    \small
    \begin{tabular}{lccc}
    \toprule
        \multirow{2}{*}{Data} &  Task Accuracy (\%)  & Prompt  \\
         & $\overbracket[0.2pt]{ \hat{\Delta_T} \; { \scriptsize  (\text{Acc}(p_c^*) \rightarrow \text{Acc}(\tilde{p}_c))}}$ & F1 (\%) \\
    \midrule
        \multirow{1}{*}{SST-2} 	& 1.0 \gr{91.9$\rightarrow$ 90.9} & 98.5 \\
    % \cmidrule(lr){1-3}
        \multirow{1}{*}{SST-5} 	& 0.9 \gr{51.4 $\rightarrow$ 50.5} &  96.1 & \\
    % \cmidrule(lr){1-3}
        \multirow{1}{*}{AGNews}	& 1.4 \gr{91.8 $\rightarrow$ 90.4} &  95.7 & \\
    % \cmidrule(lr){1-3}
        \multirow{1}{*}{Subj}	& 4.1 \gr{89.8 $\rightarrow$ 85.6}  & 100.0 & \\
    % \cmidrule(lr){1-3}
        \multirow{1}{*}{TREC}	& 0.5 \gr{88.6 $\rightarrow$ 88.1}  & 99.3 & \\
    \bottomrule 
    \end{tabular}
    \caption{
         Accuracy of solving five classification datasets, 
         unconstrained setting ($p_c^*$)
         vs.\ constrained by the projection to {\bf the true definition of tasks} ($\tilde{p}_c$) using $\gamma = 0.01$ in the objective function (Eq.\ref{eq:modified:obj}).
        %  $\hat{\Delta}$ indicates the accuracy drop (in \%) from unconstrained accuracy.
        %  Each reported score ({\em Accuracy} and {\em Prompt F1}) are the average over 62 discrete target prompts and 3 random seeds.
        %  \textbf{Overall, it is possible to achieve $\boldsymbol{\geq 94\%}$  prompt F1 with under $\boldsymbol{2\%}$ drop in accuracy.}
        Subscript $T$ in $\Delta_T$ denotes this being the case for true task definitions.
        {\bf Projecting to the true definition of a task does not help continuous prompts solve a task.}
    }
    \label{tab:results:true:definitions}
\end{table}
\begin{table}[ht]
    \centering
    \small 
    \setlength\tabcolsep{1ex}
    % \begin{tabular}{ccc}
    %     \toprule 
    %     Data & $\hat{\Delta}_t$ & $\hat{\Delta}$ \\
    %     \midrule
    %     \multirow{1}{*}{SST-2}  & 1.0 & 0.6 \\
    %     \multirow{1}{*}{SST-5} 	& 0.9 & 2.0 \\
    %     \multirow{1}{*}{AGNews}	& 1.4 & 0.8 \\
    %     \multirow{1}{*}{Subj}	& 4.1 & 1.5 \\
    %     \multirow{1}{*}{TREC}	& 0.5 & 2.3 \\
    %     \midrule 
    %     Avg & 1.6 & 1.4 \\ 
    %     \bottomrule
    % \end{tabular}
    \begin{tabular}{c|ccccc|c}
        \toprule 
             & SST-2  & SST-5 & AGNews & Subj &TREC & Avg  \\
        \midrule 
            $\hat{\Delta}_T$ & 1.0 & 0.9 & 1.4 & 4.1  & 0.5 & 1.6 \\ 
            $\hat{\Delta}$ & 0.6  & 2.0 &  0.8 & 1.5 &  2.3 & 1.4 \\  
        \bottomrule 
    \end{tabular}
    \caption{
        Task accuracy gap comparison between unconstrained prompts and those fine-tuned to project to a \emph{true} task definition ($\hat{\Delta}_T$) as reported in Table~\ref{tab:results:true:definitions}. 
        For comparison, we also show the corresponding performance gaps with \emph{irrelevant}  ($\hat{\Delta}$) from Table~\ref{tab:main_result}.
        {\bf The average performance gaps are about the same (around 1.5) for true and irrelevant target prompts}---further evidence that continuous prompts don't relate to the task being solved.
    }
    \label{tab:results:true:definitions:deltas}
\end{table}

% \section{Experiment: Projection onto True Task Definitions}

\paragraph{Projection onto true task definitions.}
In all our results so far in \S\ref{sec:empirical}, the target projected text was orthogonal to the tasks being solved. 
One might naturally wonder whether there is any benefit in projecting continuous prompts to the texts that truly describe the task being solved, i.e., a ``true'' prompt for the task.
To this end, we manually authored 5 ``true'' prompts for each of the tasks.
We then follow the exact same setup used earlier for Table~\ref{tab:main_result} to fine-tune continuous prompts $\tilde{p}_c$ for the task while projecting onto these true task definitions. 
As before, we fine-tune unconstrained prompts $p^*_c$ of the same length, without any projection requirement.

By design, $\tilde{p}_c$ can be no more effective at solving the task than the unconstrained prompt $p^*_c$ (barring suboptimal search issues), which is what we find in practice. For completeness, we report detailed results for ``true'' target prompts (analogous to Table~\ref{tab:main_result}) in Table~\ref{tab:results:true:definitions}.

More interestingly, as shown in Table~\ref{tab:results:true:definitions:deltas}, \textbf{continuous prompts that project to ``true'' target prompts are no more effective at solving the task} than continuous prompts that project to the 62 \emph{irrelevant} target prompts considered earlier (Table~\ref{tab:main_result}). Specifically, the average performance gap $\Delta$ (relative to unconstrained prompts of the same length) is about the same ($\approx 1.5\%$) for continuous prompts that map to true task definitions compared to prompts that map to irrelevant text.
This  further bolsters the waywardness hypothesis---continuous prompts don't relate to the task being solved.

% The results are summarized in Table~\ref{tab:results:true:definitions}. 
% As the result show, training continuous prompts that are projected to the true definition of the tasks they solve does not improve their quality.\ashish{Improve w.r.t.\ what? See comment in 4.2. Clearly it can't improve w.r.t.\ the optimal continuous prompt. But we can compare it to $\tilde{p}_c$ from that earlier section and report how good/bad this is compared to that. From eyeballing, the numbers would look quite good!} 
% This is in accord with the waywardness hypothesis that stipulates a disconnect between the task solved by continuous and their overall effect.  
% This is in accord with the waywardness hypothesis -- continuous prompts don't relate to the task being solved.
% The results seems to match the trend we have been seeing: we are able to get continuous prompts with very low accuracy drop and high F1 score when being mapped and compared to the target discrete prompt.

% \subsection{Why is \shortname{} happening?}
% \subsection{Intuitions Behind \shortname{}}
% \subsection{Making Sense of \shortname{}}
\section{Explaining \shortname{}}
\label{subsec:intuiton}

% \sewon{I might have misunderstood, but I am not entirely sure the connection between this section and the high-level story. I think an arbitrary discrete prompt being mapped to an infinite number of continuous prompts is quite trivial (we probably even do not need a proof, or maybe can move proofs to the appendix). But how is it connected to the fact that some of these continuous prompts are good prompts for some task A which is unrelated to the original discrete prompt, which I think is more important for Prompt Waywardness? }\tushar{Reading this section more closely, it does seem trivial (it is basically the argument that there are infinite real numbers next to any integer). It is hard to prove the last statement in Sewon's comment but maybe we can clarify that this section is mainly trying to give an intuitive sense of why waywardness could be possible.}

Here we provide intuitions behind the factors that enable \name. 
% \paragraph{Infinite mappings from continuous to discrete space. }

\paragraph{The mapping between continuous and discrete spaces is not one-to-one. }
% The mapping between the space of discrete input and that of word embeddings  (Fig.\ref{fig:setup}) 
% is \underline{not} a bijection. 
While a discrete target prompt is mapped to exactly one continuous prompt (via its embedding, Eq.\ref{eq:cproj}; cf.~Fig.\ref{fig:setup}), the reverse is not true. In fact, except for a very small fraction of unnatural or degenerate edge cases,\footnote{Such as using a non-metric distance in nearest-neighbor mapping, or mapping all of $\mathbb{R}^d$ to a single discrete prompt.}
% \sewon{Maybe we don't need `in general' as it is strictly always true?} \ashish{it's actually not always true without assumptions like those in the propositions below. E.g., consider a silly projection that maps ALL of $\mathbb{R}^d$ to one discrete target prompt. Or a projection that maps exactly the embedding of $p_d$ back to $p_d$, and all non-embeddings to one default discrete point. But these are non-natural (the point of Prop 1) or outliers (the point of Prop 2).} 
for every target discrete prompt, there are infinitely many continuous prompts that project back to it (via Eq.\ref{eq:dproj}). While simple counting-based arguments are insufficient in continuous spaces, we formally prove (Appendix~\ref{sec:appedix:not:bijection}) that this property holds for all nearest-neighbor projections under any metric distance, and broadly for all but a negligible (measure zero) portion of possible projection operators.
% \danielTwo{TODO: Ashish will massage the above paragraph to convey brief, non-trivial intuitions about the nature of this mapping. }

\begin{figure}[tb]
    \centering
    \includegraphics[trim=0.6cm 0.84cm 0cm 0.45cm, scale=0.89]{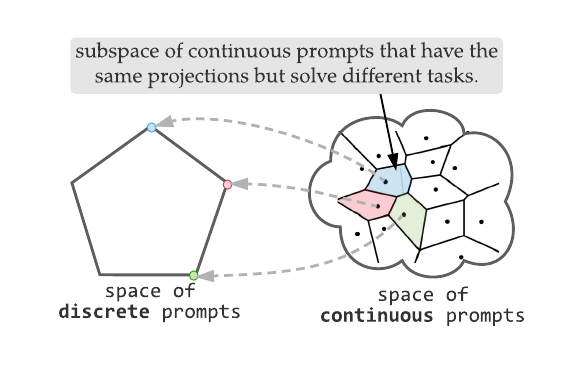}
    \caption{The projection discrete space (Eq.\ref{eq:dproj}) induces a clustering (a Voronoi diagram) of the continuous space. Each cluster has infinitely many points that get mapped to the same discrete token. 
    % \daniel{be explicit that this is for a fixed position.}
    }
    \label{fig:clustering}
\end{figure}

This intuitively suggests that there is a whole region of continuous prompts that corresponds to a fixed discrete representation (Fig.\ref{fig:clustering}). 
The remaining question is, how is this region able to have a diverse set of prompts that can solve a variety of tasks? This is addressed next.

\paragraph{Deep models give \changed{immense} expressive power to earlier layers. }
% Another factor is the multitude of layers in LMs which give a lot of power to the earlier layers and the input of the network. 
% Because of the multiplicative effect of the layers, small changes in the input can lead to massive changes in the output. 
The deeper a network is, the more expressivity it has with respect to its inputs~\cite{telgarsky2016benefits,raghu2017expressive}. 
Since continuous prompts reside just before the first layer, they enjoy a lot of expressivity. 
Therefore, no matter how narrow the regions corresponding to individual tokens are (Fig.\ref{fig:clustering}), they are extremely powerful in solving a variety of tasks. 
Previously in \S\ref{subsec:result} we provide an empirical analysis showing evidence that the effect of \shortname{} is stronger in deeper models.

\section{Implications of \name}
\label{subsec:implications}
% \daniel{
% I wonder if there is a better way to structure this section. 
% The way these are organized it seems that everything is equally important and independent. But one might argue that some of these issues are co-dependent.
% }

We discuss the implications of these findings on several inter-related lines of research. 
Note that all the following statements are valid within the boundaries of the existing architectures. Moving beyond these barriers likely requires major innovations in terms of LM architectures or how continuous prompts are optimized.

\paragraph{Faithful interpretation of continuous prompts is difficult.}
% \sewon{What about adding ``with a current best way of projection'' or replace ``unlikely'' to ``difficult''?}
% While we are not aware of a work on interpreting any continuous prompts, we hypothesis  
Given the intuitions behind and empirical support for the \shortname\thinspace hypothesis (\S\ref{subsec:intuiton}), faithful discrete interpretations of continuous prompts via common discrete projections (like nearest-neighbor projection) are unlikely to be robust based on current approaches. 
It is an open question whether there is a better way of interpreting continuous prompts with human language, or whether explaining and interpreting continuous prompts via human language is inherently impossible because they lie in completely different spaces. 
Future work may investigate more on this topic in order to improve the interpretability of prompt-based language models.

% \sewon{Feel like we need some more elaboration here..} \yejin{unlikely to be successful => unlikely to be robust, to soften the wording a bit}
% Continuous prompts cannot (and should not) be “explained” using discrete projections. 
% This is unless there will be innovations in terms of how continuous prompt optimization is performed.
% \tushar{Maybe just say "interpreting continuous prompts (trained using current approaches) in terms of.."}\daniel{rephrasing it a bit. See if it's better?}\tushar{I should have clarified my goal here. The last sentence here makes it seem that your results are narrowly focused on current prompt tuning approaches. In a way, the empirical results are based on the current approaches but I am not sure if new approaches would lead to any new results. So rather than highlighting this caveat (which is not important), just mention that your results are based on current approaches.}
% \daniel{got it. what bothers me is that "current approaches" is ambiguous. What constitutes as "current approaches"?}\tushar{True. In a way, it is unclear if this result even applies to the other "current" prompt-tunping apro, so can't eaches.}
% \sewon{How about making it more like ``The most straight-forward way of interpreting the continuous prompt (d-proj) is unfaithful'', as there might be alternative way of interpreting the continuous pomrpt?}

\paragraph{Risk of interpreting continuous prompts: concealed adversarial attacks.}
% \kyle{In a future in which proprietary model development is driven by human-driven prompt engineering, one consequence of our results is that such prompts can carry alarming adversarial behavior since the discrete projection of prompts can appear benign yet hide the potentially malicious intention of the prompt designer.}
It is not difficult to imagine a future where proprietary model development is driven by fine-tuned continuous prompts. 
In such a world, not addressing the challenges involved in discrete interpretation of continuous prompts can lead to harmful (and potentially, adversarial) consequences~\cite{slack2020fooling,wallace2021concealed}, as discussed below. 
% , they can carry adversarial behaviors and their discrete projection would not reveal their true function/intention. In fact, it can be used to justify them as harmless.  
% While the concrete application might not exist yet, since using prompts is becoming popular, our intention is to foresee any such usage and prevent its harms.

\begin{figure}[ht]
    \centering
    \includegraphics[scale=0.88,trim=0.63cm 0.68cm 0.5cm 0.79cm,clip=true]{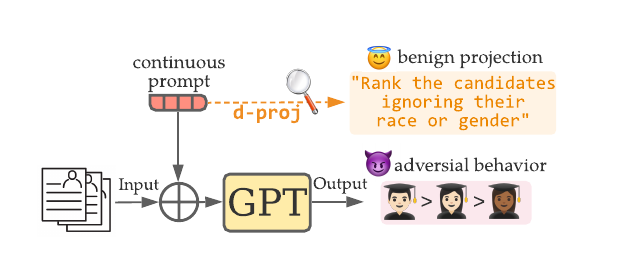}
    \caption{
    % Concealed attacks via continuous prompts. A continuous prompt may carry adversarial behaviors, even though its interpretation via projection might express an benign behavior.
    % A key implication of \shortname{} is that careless attempts to interpret continuous prompts can result in vulnerabilities against malicious attacks concealed under the guise of benign discrete representation. 
    % For example, a  continuous prompt touted as a fair resume selection model based on its discrete interpretation might be carrying egregious biases. 
    \shortname{} implies that  continuous prompts can be mapped to  seemingly innocuous descriptions while acting maliciously.
    }
    \label{fig:adversial}
\end{figure}

We consider the following scenario: a model designer comes up with a set of continuous prompts that solve a target task (e.g., ranking resumes according to each applicant's qualifications and merits). Whether intentionally or not, such prompts may maliciously target, for example,  a minority group. 
To assure their customers, the model designer uses the projection of the prompt that expresses a benign definition for the task, which does not reveal the true nature of the egregious behavior. 
% The customers might even test out the prompt on a few instances from the evaluation set. However, since the adversarial behavior targets a minority group, it might not appear harmful in evaluation sets that are not comprehensive. 
% In a way, discrete projection onto a benign definition might be the reason for a sloppy evaluation of the task prompt.
The customers might even evaluate the prompt on a few instances but not notice this harmful behavior, e.g., when it effects a minority group not in the evaluation set. In a way, the benign discrete projections may provide a false sense of security. 

\paragraph{Optimizing discrete prompts through continuous prompts can be degenerate.}
% \paragraph{Optimizing Discrete Prompts through Continuous Prompts is Futile}
% \paragraph{Continuous differentiable optimization in search of discrete human-readable prompts can lead to degenerate solutions.}
% \tushar{Optimizing for human-readable prompts via continuous differentiation ....}
Manually-written discrete prompts have many nice properties~\cite{schick2021exploiting,mishra2021reframing}, yet we do not have an efficient algorithmic way of finding them. 
One way to operationalize this is to formulate differentiable objective functions via 
LMs like GPT~\cite{radford2019language}.
% Is there a way to use continuous and differentiable optimization on top of auto-regressive LMs like GPT~\cite{radford2019language} to find human readable prompts? 
% Alas, \shortname{} can poses a challenge toward solving these optimization problems.  
% A differentiable prompt optimization can be described as an optimization in the space of embeddings $p_c \in \reals^d$ such that: 
Consider the following problem which is defined in the space of continuous embeddings $p_c \in \reals^d$: 
\begin{equation}
\label{eq:prompt:opt}
\displaystyle \max_{ p_c \in \reals^d } \,\, \overbrace{\prob{\dataset | p_c }}^{\text{utility}}  \times \overbrace{\prob{ d\text{-proj}(p_c) }}^{\text{readability}}, 
\end{equation}
This a joint optimization towards a \emph{utility} objective (the extent to which it can solve dataset $\dataset$) and a \emph{human readability} objective. According to the \shortname{} hypothesis, there are $p_c$'s that assign high mass to the utility term 
% (high $\prob{\dataset | p_c}$) 
while also mapping to human interpretable text 
% (high $\prob{ d\text{-proj}(p_c)}$) 
that is \emph{irrelevant} (or even contradictory) to the task solved by the prompt -- hence,  \emph{degenerate} solutions.
% In other words, there are many \emph{degenerate} solutions to Eq.\ref{eq:prompt:opt}, making this an uninformative objective if the ultimate goal is to find effective \emph{discrete} prompts.

The same challenge holds if this optimization objective, instead of continuous prompts, is reformulated in terms of word probabilities (e.g., similar to \citet[Sec 2.2]{kumar2021controlled}). 
% We hypothesize that this challenge holds for many reformulations of this objective. 
% For example, consider an optimization that, instead of continuous prompts, is done in the space of word probabilities (e.g., see \citet[Sec 2.2]{kumar2021controlled}). 
This is the case, since searching in the space of word probabilities is analogous to a search in embedding spaces.\footnote{
\label{ft:word:distributions}
A distribution over words $p \in [0, 1]^{V}$ corresponds to a continuous prompt $p_c = c\text{-proj}(p)$ which is a weighted combination of $V$-many basis vectors (word embeddings)
% $\sum_{i=1}^{V} p[i] E[i]$ 
 that form a linear span of $\reals^d$. 
}

In summary, \shortname{} presents a challenge for searching effective discrete prompts via continuous optimization.
% mapping an embedding space into meaningful and readable textual representation that is faithful to its output behavior. 
The recent works have used additional signals such as domain-specific constraints~\cite{qin2020backpropagation,Khot2021TextMN,qin2022cold} to alleviate these challenges. 
% several existing works have tackled variants of this problem 
% (abductive generation for language inference~\cite{qin2020backpropagation} and for sub-question generation~\cite{Khot2021TextMN}) by employing a variety of constraints as additional, domain-specific signals. 
We hope to see more design innovations 
% for further progress 
in this direction.

\paragraph{Gradients \emph{alone} are insufficient to reverse engineer a model.}
% \yejin{i like the catchy section heading, but can we make the following text extra precise by specifying gradients with respect to `what', because we don't empirically or theoretically demonstrate that gradients are insufficient for all possible objective functions}\daniel{revised the phrasing a bit}
Suppose we are given a fixed (fine-tuned or otherwise) model $M$ (e.g., an open question-answering model) and an expected output $y$ from this model (e.g., $y=$\emph{``Joe Biden''}).
Can we use gradients \changedTwo{with respect to an LM's \emph{output}} to find a semantically meaningful \emph{input} that makes a frozen model $M$ generate a particular output? 
% (without any additional assumptions on the input.)

\begin{comment}
Can we use \emph{gradients} alone to generate a semantically meaningful input question that makes the model $M$ generate this given answer? (without any additional assumptions on the input).
More formally, if $q \in [0, 1]^{L\times V}$ is a probability distribution over all questions of length $L$, are gradients with respect to the question input, $\frac{\partial M(c\text{-proj}(q)) = y}{\partial q_{lv}}$, alone informative enough to move us towards the best human readable input that is faithful to the task being solved by $M$?\tushar{What purpose does the formalization here serve since we don't use it? It would be simpler to understand if we just stopped at "Can we use \emph{gradients} alone to generate a semantically meaningful input question that makes the model $M$ generate this given answer? (without any additional assumptions on the input).}
\end{comment}

Our findings and the earlier argument about continuous differentiable optimization suggests this may not be feasible with current methods. 
To see the correspondence to \name{}, we can replace $\dataset$ in Eq.\ref{eq:prompt:opt} with
% Unfortunately, our answer is a `\emph{no}' 
% these gradients alone are probably not informative enough to for this reverse engineering problem with the best human readable inputs, 
% based on  \name{} and the arguments in the earlier paragraph. 
% This is easy to see if in Eq.\ref{eq:prompt:opt} we replace $\dataset$ 
the desired outcome $y$ and run the optimization over word distributions (cf.~Footnote~\ref{ft:word:distributions}). 
While gradients can guide
% help guide us 
towards \emph{some} input that makes $M$ produce $y$, 
their interpretation is likely unfaithful 
% input is quite likely to not be faithful 
to the task being solved by $M$. 
In the context of the above example ($M$ being a QA system), gradients might lead to inputs maximize the probability assigned to \emph{``Joe Biden''}, although this input will likely be  
neither fluent nor semantically descriptive of \emph{``Joe Biden''.}
% that are perhaps linguistically fluent but are neither proper queries nor semantically descriptive of \emph{``Joe Biden''}. 
% \sean{does this contradict the paragraph above discussing constraints-- that paragraph made it seem like it's possible with the proper regularization.}
% \daniel{revised the paragraph below. check?}
% \yejin{I think this statement is too strong. If people come up with a different objective function to optimize, e.g., bi-directional prob, then the answer might be different. or at least this paper does not empirically demonstrate that far... similarly, some folks might come up with a different projection strategies between C and D that lead to different empirical behaviors...}\ashish{+1. You are also hedging later about it, which is good. Could instead say: Unfortunately, our findings and the earlier argument about continuous differentiable optimization suggests this may not be feasible with current methods. To see the correspondence to \name{}, we can replace $\dataset$ in Eq.\ref{eq:prompt:opt} with ...}
% \daniel{Right, I reworded the phrasing as per Ashish's suggestion to soften the claim.} \yejin{the revision looks good to me!}

Nevertheless, as noted earlier, gradients are still useful when they are applied using domain-specific constraints. For example, one can find local (word-level) perturbations that lead to a certain adversarial outcome, if the perturbations are restricted to well-defined semantic categories (e.g., ``blue'' can be perturbed to any other color name)~\cite{sha2020gradient,guo2021gradient,yuan2021bridge}.

\paragraph{Continuous prompt tuning does not necessitate task-specific initialization.}
Recent works on continuous prompt-tuning have shown the effectiveness of initialization from embeddings of \emph{random} common words~\cite{lester2021power,min2021noisy}, despite these words being irrelevant to the task solved by these prompts. 
% In other words, initialization from the embeddings of many irrelevant words 
% It is surprising that many of these random words are irrelevant to the task solved by these prompts. 
This, however, makes sense given the observations made in this work regarding the existence of effective prompts around word embeddings. 
% subspaces. \ashish{perhaps `around arbitrary word embeddings'? the `arbitrary' would emphasize that it's not task-specific. (not sure what `around XYZ subspaces' means)}
% Several recent works on continuous prompt-tuning have used initialization from embeddings of popular words~\cite{lester2021power,min2021noisy}. 
% Given the observations in this work regarding the feasibility of finding strong continuous prompts in different sub-spaces of the continuous space, we suspect that such initialization techniques probably remain good enough. 
 %\tushar{Instead of just agreeing, we could even further propose that "using popular words is probably not even necessary; there is an optimal prompt next to any discrete token sequence"}\daniel{I am a bit worried to go that far. If you look at \cite{lester2021power} (Fig3-b) there is a clear difference between random initialization and initialization from popular vocab. }\tushar{Isn't the random completely random? What about random tokens? I do agree that we don't have proof for this so we might be going too far. What if we just say  that using popular words \emph{may} not be necessary as long as the prompt vectors correspond to some words.}\daniel{I see. Rephrased it a bit. Better?}

% \section{Closing Remarks}\sewon{Nitpicking, but why not `Conclusion'? Closing Remarks for a paper seems a bit odd to me...}
\section{Conclusion}
The prompting literature has seen many parallel developments \changed{around continuous and discrete prompts, as efficient alternatives to fine-tuning models with tens of millions of parameters.} 
Our work introduced the \name{} hypothesis, which expresses a surprising disconnect between continuous and discrete prompts: \changed{given a downstream task,}
for \emph{any} discrete target prompt $p_d$, there exists a continuous prompt that projects to $p_d$ while achieving strong performance on \changed{the task.}
We provided empirical evidence for this hypothesis, studied various parameters around it, and ended with several implications of this hypothesis.
% there exists strong continuous prompts for solving tasks in a way that their discrete projections have nothing to do with the task they solve. 
%  \yejin{this statement reads overly strong due to "nearly perfect"...}
%  \ashish{Related to this: what is ``showed a stronger claim'' referring to? the Hypothesis? or the empirical support of it? If the former, avoid "showed" and "claim. Could instead say proposed a general hypothesis:}
% \daniel{reworded}

% \paragraph{Extent of the findings.}
While our experiments are done on the GPT family, we expect 
% many of 
our findings to apply to a broader set of architectures that, in one way or another, use similar mechanisms for mapping discrete elements to continuous representations 
% (Eq.\ref{eq:cproj})  
and vice versa. 
Similarly, \changed{while} our projection to the discrete space (Eq.\ref{eq:dproj}) is a popular operator in the field 
(cf.~Footnote~\ref{footnote1:operator:generality}),
% choice that is used in the last later\tushar{something is off} of many sequence-to-sequence LMs~\cite{lewis2020bart,raffel2020exploring}. 
the intuition explained in Propositions~\ref{prop:nnbr-projections} and~\ref{prop:infinite-to-one} \changed{of the Appendix} \changed{suggests similar behavior} for a broad class of projection operators.  

\name{} \changed{identifies challenges} for future progress on \changed{algorithmic methods for the discovery} of human readable
prompts that are faithful to the task they solve.
% \ashish{More critical to point out than human readable (which you can achieve with an LM loss) is \emph{faithful} / semantically meaningful.}
% automatic general-purpose discovery of readable discrete prompts. 
% However, it is conceivable that we hope that architectural innovations can overcome such challenges.  
% For example, learned projections for mapping continuous vectors to discrete space might be effective for some problems, 
We hope the observations made in this work 
motivate architectural innovations that overcome such challenges and guide future steps in the prompting literature.

% \daniel{
% Q: Have anyone used projections to explain continuous prompts?
% Not that that we know. But we want to foresee (and prevent) any such usages.  \\ 
% }
% \daniel{
% limits of this study: the projection (update: not a limit), the model (gpt2 family), etc. 
% }

\section*{Acknowledgment}
The authors are thankful to Lisa Li, Nicholas Lourie, and Vered Shwartz for helpful discussions, and the Beaker team at AI2 for their support with the experiments. 
This work was supported in part by DARPA MCS program through NIWC Pacific (N66001-19-2-4031),
DARPA SemaFor program, and Google Cloud Compute.
% and the Allen Institute for AI.  % dropping as there are full-time AI2 folks on the author list

% Entries for the entire Anthology, followed by custom entries
\bibliography{ref}

\providecommand{\CNFX}[1]{{\em{\textrm{#1}}}}
\begin{thebibliography}{38}
\expandafter\ifx\csname natexlab\endcsname\relax\def\natexlab#1{#1}\fi

\bibitem[{Brown et~al.(2020)Brown, Mann, Ryder, Subbiah, Kaplan, Dhariwal,
  Neelakantan, Shyam, Sastry, Askell et~al.}]{brown2020language}
Tom~B Brown, Benjamin Mann, Nick Ryder, Melanie Subbiah, Jared Kaplan, Prafulla
  Dhariwal, Arvind Neelakantan, Pranav Shyam, Girish Sastry, Amanda Askell,
  et~al. 2020.
\newblock Language models are few-shot learners.
\newblock \emph{arXiv preprint arXiv:2005.14165}.

\bibitem[{Dathathri et~al.(2019)Dathathri, Madotto, Lan, Hung, Frank, Molino,
  Yosinski, and Liu}]{dathathri2019plug}
Sumanth Dathathri, Andrea Madotto, Janice Lan, Jane Hung, Eric Frank, Piero
  Molino, Jason Yosinski, and Rosanne Liu. 2019.
\newblock Plug and play language models: A simple approach to controlled text
  generation.
\newblock In \emph{Proceedings of \CNFX{ICLR}}.

\bibitem[{Gao et~al.(2021)Gao, Fisch, and Chen}]{gao2021making}
Tianyu Gao, Adam Fisch, and Danqi Chen. 2021.
\newblock Making pre-trained language models better few-shot learners.
\newblock In \emph{Proceedings of \CNFX{ACL}}.

\bibitem[{Guo et~al.(2021)Guo, Sablayrolles, J{\'e}gou, and
  Kiela}]{guo2021gradient}
Chuan Guo, Alexandre Sablayrolles, Herv{\'e} J{\'e}gou, and Douwe Kiela. 2021.
\newblock Gradient-based adversarial attacks against text transformers.
\newblock In \emph{Proceedings of \CNFX{EMNLP}}.

\bibitem[{Hashimoto et~al.(2016)Hashimoto, Alvarez-Melis, and
  Jaakkola}]{hashimoto2016word}
Tatsunori~B Hashimoto, David Alvarez-Melis, and Tommi~S Jaakkola. 2016.
\newblock Word embeddings as metric recovery in semantic spaces.
\newblock \emph{\CNFX{TACL}}, 4:273--286.

\bibitem[{Jiang et~al.(2020)Jiang, Xu, Araki, and Neubig}]{jiang2020can}
Zhengbao Jiang, Frank~F Xu, Jun Araki, and Graham Neubig. 2020.
\newblock How can we know what language models know?
\newblock \emph{\CNFX{TACL}}, 8:423--438.

\bibitem[{Khot et~al.(2021)Khot, Khashabi, Richardson, Clark, and
  Sabharwal}]{Khot2021TextMN}
Tushar Khot, Daniel Khashabi, Kyle Richardson, Peter Clark, and Ashish
  Sabharwal. 2021.
\newblock {T}ext {M}odular {N}etworks: Learning to decompose tasks in the
  language of existing models.
\newblock \emph{Proceedings of \CNFX{NAACL}}, page 1264–1279.

\bibitem[{Kumar et~al.(2021)Kumar, Malmi, Severyn, and
  Tsvetkov}]{kumar2021controlled}
Sachin Kumar, Eric Malmi, Aliaksei Severyn, and Yulia Tsvetkov. 2021.
\newblock Controlled text generation as continuous optimization with multiple
  constraints.
\newblock \emph{Proceedings of \CNFX{NeurIPS}}, 34.

\bibitem[{Le~Scao and Rush(2021)}]{le2021many}
Teven Le~Scao and Alexander~M Rush. 2021.
\newblock How many data points is a prompt worth?
\newblock In \emph{Proceedings of \CNFX{NAACL}}, pages 2627--2636.

\bibitem[{Lester et~al.(2021)Lester, Al-Rfou, and Constant}]{lester2021power}
Brian Lester, Rami Al-Rfou, and Noah Constant. 2021.
\newblock The power of scale for parameter-efficient prompt tuning.
\newblock In \emph{Proceedings of \CNFX{EMNLP}}.

\bibitem[{Li and Liang(2021)}]{li2021prefix}
Xiang~Lisa Li and Percy Liang. 2021.
\newblock Prefix-tuning: Optimizing continuous prompts for generation.
\newblock In \emph{Proceedings of \CNFX{ACL}}.

\bibitem[{Mikolov et~al.(2013)Mikolov, Sutskever, Chen, Corrado, and
  Dean}]{mikolov2013distributed}
Tomas Mikolov, Ilya Sutskever, Kai Chen, Greg~S Corrado, and Jeff Dean. 2013.
\newblock Distributed representations of words and phrases and their
  compositionality.
\newblock In \emph{Proceedings of \CNFX{NeurIPS}}, pages 3111--3119.

\bibitem[{Min et~al.(2022)Min, Lewis, Hajishirzi, and
  Zettlemoyer}]{min2021noisy}
Sewon Min, Mike Lewis, Hannaneh Hajishirzi, and Luke Zettlemoyer. 2022.
\newblock Noisy channel language model prompting for few-shot text
  classification.
\newblock In \emph{Proceedings of \CNFX{ACL}}.

\bibitem[{Mishra et~al.(2022{\natexlab{a}})Mishra, Khashabi, Baral, Choi, and
  Hajishirzi}]{mishra2021reframing}
Swaroop Mishra, Daniel Khashabi, Chitta Baral, Yejin Choi, and Hannaneh
  Hajishirzi. 2022{\natexlab{a}}.
\newblock Reframing instructional prompts to {GPTk}'s language.
\newblock In \emph{Proceedings of \CNFX{ACL} - Findings}.

\bibitem[{Mishra et~al.(2022{\natexlab{b}})Mishra, Khashabi, Baral, and
  Hajishirzi}]{mishra2021cross}
Swaroop Mishra, Daniel Khashabi, Chitta Baral, and Hannaneh Hajishirzi.
  2022{\natexlab{b}}.
\newblock Cross-task generalization via natural language crowdsourcing
  instructions.
\newblock In \emph{Proceedings of \CNFX{ACL}}.

\bibitem[{Pang and Lee(2004)}]{pang2004sentimental}
Bo~Pang and Lillian Lee. 2004.
\newblock A sentimental education: Sentiment analysis using subjectivity
  summarization based on minimum cuts.
\newblock In \emph{Proceedings of \CNFX{ACL}}.

\bibitem[{Petroni et~al.(2019)Petroni, Rockt{\"a}schel, Riedel, Lewis, Bakhtin,
  Wu, and Miller}]{petroni2019language}
Fabio Petroni, Tim Rockt{\"a}schel, Sebastian Riedel, Patrick Lewis, Anton
  Bakhtin, Yuxiang Wu, and Alexander Miller. 2019.
\newblock Language models as knowledge bases?
\newblock In \emph{Proceedings of \CNFX{EMNLP}}.

\bibitem[{Qin and Eisner(2021)}]{qin2021learning}
Guanghui Qin and Jason Eisner. 2021.
\newblock Learning how to ask: Querying lms with mixtures of soft prompts.
\newblock In \emph{Proceedings of \CNFX{NAACL}}.

\bibitem[{Qin et~al.(2020)Qin, Shwartz, West, Bhagavatula, Hwang, Le~Bras,
  Bosselut, and Choi}]{qin2020backpropagation}
Lianhui Qin, Vered Shwartz, Peter West, Chandra Bhagavatula, Jena~D Hwang,
  Ronan Le~Bras, Antoine Bosselut, and Yejin Choi. 2020.
\newblock Backpropagation-based decoding for unsupervised counterfactual and
  abductive reasoning.
\newblock In \emph{Proceedings of \CNFX{EMNLP}}.

\bibitem[{Qin et~al.(2022)Qin, Welleck, Khashabi, and Choi}]{qin2022cold}
Lianhui Qin, Sean Welleck, Daniel Khashabi, and Yejin Choi. 2022.
\newblock {COLD} decoding: Energy-based constrained text generation with
  {L}angevin dynamics.
\newblock \emph{arXiv preprint arXiv:2202.11705}.

\bibitem[{Radford et~al.(2019)Radford, Wu, Child, Luan, Amodei, Sutskever
  et~al.}]{radford2019language}
Alec Radford, Jeffrey Wu, Rewon Child, David Luan, Dario Amodei, Ilya
  Sutskever, et~al. 2019.
\newblock Language models are unsupervised multitask learners.
\newblock \emph{OpenAI blog}, 1(8):9.

\bibitem[{Raffel et~al.(2020)Raffel, Shazeer, Roberts, Lee, Narang, Matena,
  Zhou, Li, and Liu}]{raffel2020exploring}
Colin Raffel, Noam Shazeer, Adam Roberts, Katherine Lee, Sharan Narang, Michael
  Matena, Yanqi Zhou, Wei Li, and Peter~J Liu. 2020.
\newblock Exploring the limits of transfer learning with a unified text-to-text
  transformer.
\newblock \emph{\CNFX{JMLR}}, 21:1--67.

\bibitem[{Raghu et~al.(2017)Raghu, Poole, Kleinberg, Ganguli, and
  Sohl-Dickstein}]{raghu2017expressive}
Maithra Raghu, Ben Poole, Jon Kleinberg, Surya Ganguli, and Jascha
  Sohl-Dickstein. 2017.
\newblock On the expressive power of deep neural networks.
\newblock In \emph{Proceedings of \CNFX{ICML}}, pages 2847--2854.

\bibitem[{Rajpurkar et~al.(2016)Rajpurkar, Zhang, Lopyrev, and
  Liang}]{rajpurkar2016squad}
Pranav Rajpurkar, Jian Zhang, Konstantin Lopyrev, and Percy Liang. 2016.
\newblock {SQuAD}: 100,000+ questions for machine comprehension of text.
\newblock In \emph{Proceedings of \CNFX{EMNLP}}, pages 2383--2392.

\bibitem[{Reynolds and McDonell(2021)}]{reynolds2021prompt}
Laria Reynolds and Kyle McDonell. 2021.
\newblock Prompt programming for large language models: Beyond the few-shot
  paradigm.
\newblock In \emph{Proceedings of \CNFX{CHI}}.

\bibitem[{Schick and Sch{\"u}tze(2021)}]{schick2021exploiting}
Timo Schick and Hinrich Sch{\"u}tze. 2021.
\newblock Exploiting cloze-questions for few-shot text classification and
  natural language inference.
\newblock In \emph{Proceedings of \CNFX{EACL}}, pages 255--269.

\bibitem[{Sha(2020)}]{sha2020gradient}
Lei Sha. 2020.
\newblock Gradient-guided unsupervised lexically constrained text generation.
\newblock In \emph{Proceedings of \CNFX{EMNLP}}, pages 8692--8703.

\bibitem[{Shin et~al.(2020)Shin, Razeghi, Logan~IV, Wallace, and
  Singh}]{shin2020eliciting}
Taylor Shin, Yasaman Razeghi, Robert~L Logan~IV, Eric Wallace, and Sameer
  Singh. 2020.
\newblock Eliciting knowledge from language models using automatically
  generated prompts.
\newblock In \emph{Proceedings of \CNFX{EMNLP}}.

\bibitem[{Slack et~al.(2020)Slack, Hilgard, Jia, Singh, and
  Lakkaraju}]{slack2020fooling}
Dylan Slack, Sophie Hilgard, Emily Jia, Sameer Singh, and Himabindu Lakkaraju.
  2020.
\newblock Fooling lime and shap: Adversarial attacks on post hoc explanation
  methods.
\newblock In \emph{Proceedings of the AAAI/ACM Conference on AI, Ethics, and
  Society}, pages 180--186.

\bibitem[{Socher et~al.(2013)Socher, Perelygin, Wu, Chuang, Manning, Ng, and
  Potts}]{socher2013recursive}
Richard Socher, Alex Perelygin, Jean Wu, Jason Chuang, Christopher~D Manning,
  Andrew~Y Ng, and Christopher Potts. 2013.
\newblock Recursive deep models for semantic compositionality over a sentiment
  treebank.
\newblock In \emph{Proceedings of \CNFX{EMNLP}}.

\bibitem[{Telgarsky(2016)}]{telgarsky2016benefits}
Matus Telgarsky. 2016.
\newblock Benefits of depth in neural networks.
\newblock In \emph{Proceedings of \CNFX{COLT}}, pages 1517--1539.

\bibitem[{Voorhees and Tice(2000)}]{voorhees2000building}
Ellen~M Voorhees and Dawn~M Tice. 2000.
\newblock Building a question answering test collection.
\newblock In \emph{Proceedings of \CNFX{SIGIR}}.

\bibitem[{Wallace et~al.(2021)Wallace, Zhao, Feng, and
  Singh}]{wallace2021concealed}
Eric Wallace, Tony Zhao, Shi Feng, and Sameer Singh. 2021.
\newblock Concealed data poisoning attacks on {NLP} models.
\newblock In \emph{Proceedings of \CNFX{NAACL}}.

\bibitem[{Wang and Komatsuzaki(2021)}]{wang2021gpt}
Ben Wang and Aran Komatsuzaki. 2021.
\newblock {GPT-J-6B: A 6 Billion Parameter Autoregressive Language Model}.
\newblock \url{https://github.com/kingoflolz/mesh-transformer-jax}.

\bibitem[{Yuan et~al.(2021)Yuan, Zhang, Chen, and Wei}]{yuan2021bridge}
Lifan Yuan, Yichi Zhang, Yangyi Chen, and Wei Wei. 2021.
\newblock Bridge the gap between {CV} and {NLP}! a gradient-based textual
  adversarial attack framework.
\newblock \emph{arXiv preprint arXiv:2110.15317}.

\bibitem[{Zhang et~al.(2015)Zhang, Zhao, and LeCun}]{zhang2015character}
Xiang Zhang, Junbo Zhao, and Yann LeCun. 2015.
\newblock Character-level convolutional networks for text classification.
\newblock In \emph{Proceedings of \CNFX{NeurIPS}}.

\bibitem[{Zhong et~al.(2021)Zhong, Friedman, and Chen}]{zhong2021factual}
Zexuan Zhong, Dan Friedman, and Danqi Chen. 2021.
\newblock Factual probing is [mask]: Learning vs. learning to recall.
\newblock In \emph{Proceedings of \CNFX{NAACL}}, pages 5017--5033.

\bibitem[{Zhou et~al.(2021)Zhou, Yang, Loy, and Liu}]{zhou2021learning}
Kaiyang Zhou, Jingkang Yang, Chen~Change Loy, and Ziwei Liu. 2021.
\newblock Learning to prompt for vision-language models.
\newblock \emph{arXiv preprint arXiv:2109.01134}.

\end{thebibliography}
\bibliographystyle{acl_natbib}

\clearpage

\appendix

{\Large \bf Supplementary Material }

\section{Additional Experimental Details}
Here we include several experimental details (\S\ref{sec:empirical}) that did not fit in the main text.
For the experiments we used A100 GPUs with 40G memory. 
In terms of the time GPU time of the experiments, each round of training and inference for each seed took about around 6 min. 
Therefore, the total GPU hours for our main experiment (Table~\ref{tab:main_result}) adds up to 93 hours (6 mins $\times$ 3 seeds $\times$ 5 datasets $\times$ 62 prompts = 5580 mins).

\section{The mapping between continuous and discrete space is not one-to-one}
\label{sec:appedix:not:bijection}

% \paragraph{The mapping between continuous and discrete space is not one-to-one. }
As argued in \S\ref{subsec:intuiton}, the mapping between the space of discrete input and that of word embeddings  (Fig.\ref{fig:setup}) 
is \underline{not} a bijection. 
While a discrete target prompt is mapped to exactly one continuous prompt (via its embedding, Eq.\ref{eq:cproj}), the reverse is not true: except for some unnatural or rare cases (as formalized in the following propositions) 
% \sewon{Maybe we don't need `in general' as it is strictly always true?} \ashish{it's actually not always true without assumptions like those in the propositions below. E.g., consider a silly projection that maps ALL of $\mathbb{R}^d$ to one discrete target prompt. Or a projection that maps exactly the embedding of $p_d$ back to $p_d$, and all non-embeddings to one default discrete point. But these are non-natural (the point of Prop 1) or outliers (the point of Prop 2).} 
there are infinitely many continuous prompts that project back to a fixed discrete target prompt (via Eq.\ref{eq:dproj}).

Nearest-neighbor projections are arguably natural, computationally efficient, and useful in practice. Although we have considered them in the Euclidean space so far, they can be defined for an \emph{arbitrary} distance metric\footnote{\url{https://en.wikipedia.org/wiki/Metric_(mathematics)}} $m$ on $\mathbb{R}^d$. As before, consider an embedding of a lexicon of size $V$ into $\mathbb{R}^d$ and the corresponding one-hot vectors in $\{0,1\}^V$. We call $d\text{-proj}$ a \emph{nearest-neighbor projection operator w.r.t.\ $m$} if it maps each $x \in \mathbb{R}^d$  to the one-hot vector in $\{0,1\}^V$ that corresponds to the lexicon item whose embedding is closest to $x$ under metric $m$ (breaking ties arbitrarily).

\begin{proposition}
\label{prop:nnbr-projections}
    Every nearest-neighbor projection operator, under any metric, maps infinitely many elements of $\mathbb{R}^d$, forming one or more continuous subspaces, to every one-hot vector in $\{0,1\}^V$.
\end{proposition}

A proof is included in Appendix~\ref{sec:proofs}. In effect, the projection operators induce a clustering of the space of continuous prompts $\mathbb{R}^{d\times L}$ into regions that have the same discrete projection (Fig.\ref{fig:clustering}).

The infinite-to-one mapping aspect is not limited to
% a particular choice of projection operator (Eq.\ref{eq:dproj}) or even to 
the class of nearest-neighbor projection operators. It is rather an inherent property of the interaction between continuous and discrete spaces, and holds for a broader family consisting of all but a negligible portion of possible projection operators:
% The following result states that the aforementioned result hold for a much broader family of projection functions. 
% \daniel{hanna was suggesting maybe we should change this to an "axiom" or just say it in the text in the passing, if we feel that our proof is not rigorous enough.}
% \begin{lemma}
% \label{lemma:1}
% For any projection operator $d\text{-proj}$ that maps $\mathbb{R}^d$ into one-hot vectors in $\{0, 1\}^V$, there are infinite continuous prompts that can be projected onto a fixed discrete target prompt. 
% \end{lemma}
% \begin{proof}
% Any $d$-projection operator partitions $\mathbb{R}^d$ (an infinite space) into $V$ partitions. 
% Therefore, based on the pigeonhole principle there always exists at least a portioned region with infinite points. \tushar{Should we say something about degenerate projections where infinite points are mapped to a special vocabulary item and the remaining V-1 items only have a single point? We seem to be claiming that every discrete target prompt will have infinite continuous prompts. But the pigeonhole principle only proves that at least one discrete prompt will have infinite continuous prompts.}
% \end{proof}

\begin{proposition}
\label{prop:infinite-to-one}
    Let $\mathbb{D}$ denote the space of all projection operators that map $\mathbb{R}^d$ to one-hot vectors in $\{0,1\}^V$. Let $d\text{-proj}$ be a random projection drawn uniformly from $\mathbb{D}$. Then, with probability 1, $d\text{-proj}$ maps infinite elements of $\mathbb{R}^d$ to every one-hot vector in $\{0,1\}^V$.
\end{proposition}

\subsection{Proofs}
\label{sec:proofs}

\begin{proofnamed}[Proof of Prop.~\ref{prop:nnbr-projections}]
Let $c_i \in \mathbb{R}^d$ for $i \in \{1, \ldots, V\}$ be fixed vectors (denoting the embedding of words in a lexicon of size $V$). Let $e_i \in \{0,1\}^V$ denote the one-hot vector with 1 in the $i$-th position and 0 elsewhere. Since $d\text{-proj}$ is a \emph{nearest-neighbor projection operator w.r.t.\ $m$}, by definition it maps $x \in \mathbb{R}^d$ to $e_i$ whenever $x$ is closest to $c_i$, i.e., $i = \argmin_{j} m(x, c_j)$ (breaking ties arbitrarily).

Let $S_i \subseteq \mathbb{R}^d$ denote the pre-image of $e_i$, i.e., the elements that the nearest-neighbor projection $d\text{-proj}$ maps to the $i$-th one-hot vector. By definition, $c_i \in S_i$. Let $d' = \min_j m(c_i, c_j) > 0$ denote the distance of $c_i$ to the nearest $c_j$ w.r.t.\ the metric $m$. Consider the subspace $C_i = \{x \mid m(x, c_i) < d'/2\}$. By design, we have $C_i \subseteq S_i$. Further, moving $x$ by some small distance $\epsilon$ (w.r.t.\ $m$) to another point $x'$ changes its distance to $c_i$ only by at most $\epsilon$ (by the triangle inequality property of $m$). This implies that if $\epsilon$ is chosen to be small enough such that $m(x, c_i) + \epsilon < d'/2$, then $x'$ must also be in $C_i$. In other words, if $x \in C_i$, then, for a small enough $\epsilon$, the entire $\epsilon$-neighborhood of $x$ is also in $C_i$. It follows that $C_i$ is an open subset of $R^d$ and thus contains infinitely many elements forming a continuous subspace. Hence $S_i$, which contains $C_i$,e.g. also has infinite elements in one or more continuous subspaces.
% INCOMPLETE PROOF, NEED TO FLUSH OUT OR ALTER: Lastly, if $x \in S_i$, the fact that $m$ is a metric implies that there is a path in $\mathbb{R}^d$ connecting $x$ with $c_i$ such that all elements along this path are closest to $c_i$ and are hence in $S_i$. \tushar{The previous statement is based on what property of the metric function?}\ashish{don't have this sorted out. Will comment it out and soften the prop statement for now.} Thus, we have that $S_i$ is a single continuous subspace with infinitely many elements.
% DOESN'T WORK EXCEPT FOR EUCLIDEAN DISTANCE (e.g., doesn't work for Manhattan distance) : Lastly, if $x, y \in S_i$ and $z$ is any element in-between the two, i.e., $z = ax + (1-a)y$ for some $a \in [0,1]$, then $z \in S_i$. This follows from the assumption that $m$ is a metric, which implies that if the nearest neighbor switches from $c_i$ to some other $c_j$ while traveling along this line connecting $x$ to $y$, then it cannot switch back to $c_i$, which contradicts the assumption that $y$ is closest to $c_i$. Hence $S_i$ must be a single, continuous space
% TUSHAR: Okay. I was also confused about this for a bit. How important is it for it to be continuous?
\end{proofnamed}

\begin{proofnamed}[Proof of Prop.~\ref{prop:infinite-to-one}]
For simplicity, assume $V = 2$. A projection operator $d\text{-proj} \in \mathbb{D}$ can then be fully characterized by the subset $S \subseteq \mathrm{R}^d$ that it maps to any one arbitrarily chosen one-hot vector. Choosing $d\text{-proj}$ uniformly at random from $\mathbb{D}$ thus amounts to choosing the subset $S$ uniformly at random from $\mathbb{R}^d$. We show that the probability of choosing an $S$ such that $|S|$ is finite, is 0. (The same argument applies to $|\mathbb{R} \setminus S|$ being finite.)

To see this, let $\mathbb{S}_i$ denote the set of all (finite) subsets of $\mathbb{R}^d$ that have size exactly $i$. First, observe that the probability of choosing an $S$ that lies in $\mathbb{S}_0 \cup \mathbb{S}_1$ (i.e., a subset of $\mathbb{R}^d$ that has at most 1 element) is 0; this is a degenerate case in the underlying continuous probability space. Second, for any $i \geq 2$, $\mathbb{S}_i$ has the same ``size'' (in the measure theoretic sense) as $\mathbb{S}_1$, because one can construct an injective map from either one to the other---which follows from the fact that they both have the same \emph{cardinality} as the set $\mathbb{R}$.\footnote{This can be proved using the rules of cardinal multiplication applied to $\mathbb{S}_i$ viewed as (a subset of) the Cartesian product of $\mathbb{S}_1$ with itself, $i$ times.} Lastly, the space $\mathbb{S}$ of all finite subsets of $\mathbb{R}^d$ is the countable union $\cup_i \mathbb{S}_i$ of disjoint sets. Therefore, $\Pr[S \in \mathbb{S}] = \sum_i \Pr[S \in \mathbb{S}_i] = 0$.
\end{proofnamed}

\end{document}